\documentclass{article}
\usepackage{times}
\usepackage{epsfig}
\usepackage{graphicx}
\usepackage{amsmath}
\usepackage{amssymb}
\usepackage{authblk}

\usepackage[sort]{natbib}
\bibpunct{[}{]}{,}{n}{}{}
\usepackage[pagebackref=false,breaklinks=true,letterpaper=true,hidelinks=true]{hyperref}

\DeclareMathOperator*{\argmin}{argmin}

\usepackage{url}
\usepackage{color,soul}
\usepackage{tikz}
\usetikzlibrary{matrix,positioning}

\begin{document}

\title{A Study of Actor and Action Semantic Retention in Video Supervoxel Segmentation}

\author{Chenliang Xu\textsuperscript{\textasteriskcentered 1},
Richard F. Doell\textsuperscript{\textasteriskcentered 1},
Stephen Jos\'{e} Hanson\textsuperscript{\dag},
Catherine~Hanson\textsuperscript{\dag},
and Jason J. Corso\textsuperscript{\textasteriskcentered}\\
\textsuperscript{\textasteriskcentered}Department of Computer Science and Engineering\\
SUNY at Buffalo, Buffalo, NY 14260\\
\{chenlian,rfdoell,jcorso\}@buffalo.edu\\
\smallskip
\textsuperscript{\dag}Department of Psychology and Rutgers Brain Imaging Center\\
Rutgers University, Newark, NJ 07102\\
\{jose,cat\}@psychology.rutgers.edu
}
\date{\vspace{-5ex}}

\footnotetext[1]{Chenliang Xu and Richard F. Doell contributed equally to this paper.\\
This article is in review at the International Journal of Semantic Computing.}

\maketitle

\begin{abstract}
Existing methods in the semantic computer vision community seem unable to deal with the explosion and richness of modern, open-source and social video content. Although sophisticated methods such as object detection or bag-of-words models have been well studied, they typically operate on low level features and ultimately suffer from either scalability issues or a lack of semantic meaning. On the other hand, video supervoxel segmentation has recently been established and applied to large scale data processing, which potentially serves as an intermediate representation to high level video semantic extraction. The supervoxels are rich decompositions of the video content: they capture object shape and motion well. However, it is not yet known if the supervoxel segmentation retains the semantics of the underlying video content. In this paper, we conduct a systematic study of how well the actor and action semantics are retained in video supervoxel segmentation. Our study has human observers watching supervoxel segmentation videos and trying to discriminate both actor (human or animal) and action (one of eight everyday actions). We gather and analyze a large set of 640 human perceptions over 96 videos in 3 different supervoxel scales. Furthermore, we conduct machine recognition experiments on a feature defined on supervoxel segmentation, called supervoxel shape context, which is inspired by the higher order processes in human perception. Our ultimate findings suggest that a significant amount of semantics have been well retained in the video supervoxel segmentation and can be used for further video analysis.
\end{abstract}
\smallskip
{\bf Keywords:} semantic retention; computer vision; video supervoxel segmentation; action recognition.

%-------------------------------------------------------
%-------------------------------------------------------
\section{Introduction}
\label{sec:introduction}
%-------------------------------------------------------
%-------------------------------------------------------

% Important of Video Anlysis.
We are drowning in video content---YouTube, for example, receives 72 hours of video uploaded every minute. In many applications, there is so much video content that a sufficient supply of human observers to manually tag or annotate the videos is unavailable. Furthermore, it is widely known that the titles and tags on the social media sites like Flickr and YouTube are noisy and semantically ambiguous \cite{TaYaHoACMMM2009}. Automatic methods are needed to index and catalog the salient content in these videos in a manner that retains the semantics of the content to facilitate subsequent search and ontology learning applications.

% Limitation of SOA Vision Tech on Video.
However, despite recent advances in computer vision, such as the deformable parts model for object detection \cite{FeGiMcPAMI2010}, the scalability as the semantic space grows remains a challenge. For example, the state of the art methods on the ImageNet Large Scale Visual Recognition Challenge \cite{ILSVRC2011} have accuracies near 20\% \cite{DeBeLiECCV2010} and a recent work achieves a mean average precision of 0.16 on a 100,000 class detection problem \cite{DeRuSeCVPR2013}, which is the largest such multi-class detection model to date. To compound this difficulty, these advances are primarily on images and not videos. Methods in video analysis, in contrast, still primarily rely on low-level features \cite{TaAlYuCVPR2012}, such as space-time interest points \cite{LaIJCV2005}, histograms of oriented 3D gradients \cite{KlMaScBMVC2008}, or dense trajectories \cite{WaKlScCVPR2011}. These low-level methods cannot guarantee retention of any semantic information and subsequent indices likewise may struggle to mirror human visual semantics. More recently, a high-level video feature, called Action Bank \cite{SaCoCVPR2012}, explicitly represents a video by embedding it in a space spanned by a set, or bank, of different individual actions. Although some semantic transfer is plausible with this Action Bank, it is computationally intensive and struggles to scale with the size of the semantic space; it is also limited in its ability to deduce viewpoint invariant actions. 

\begin{figure}[t!]
  \centering
  \includegraphics[width=\textwidth]{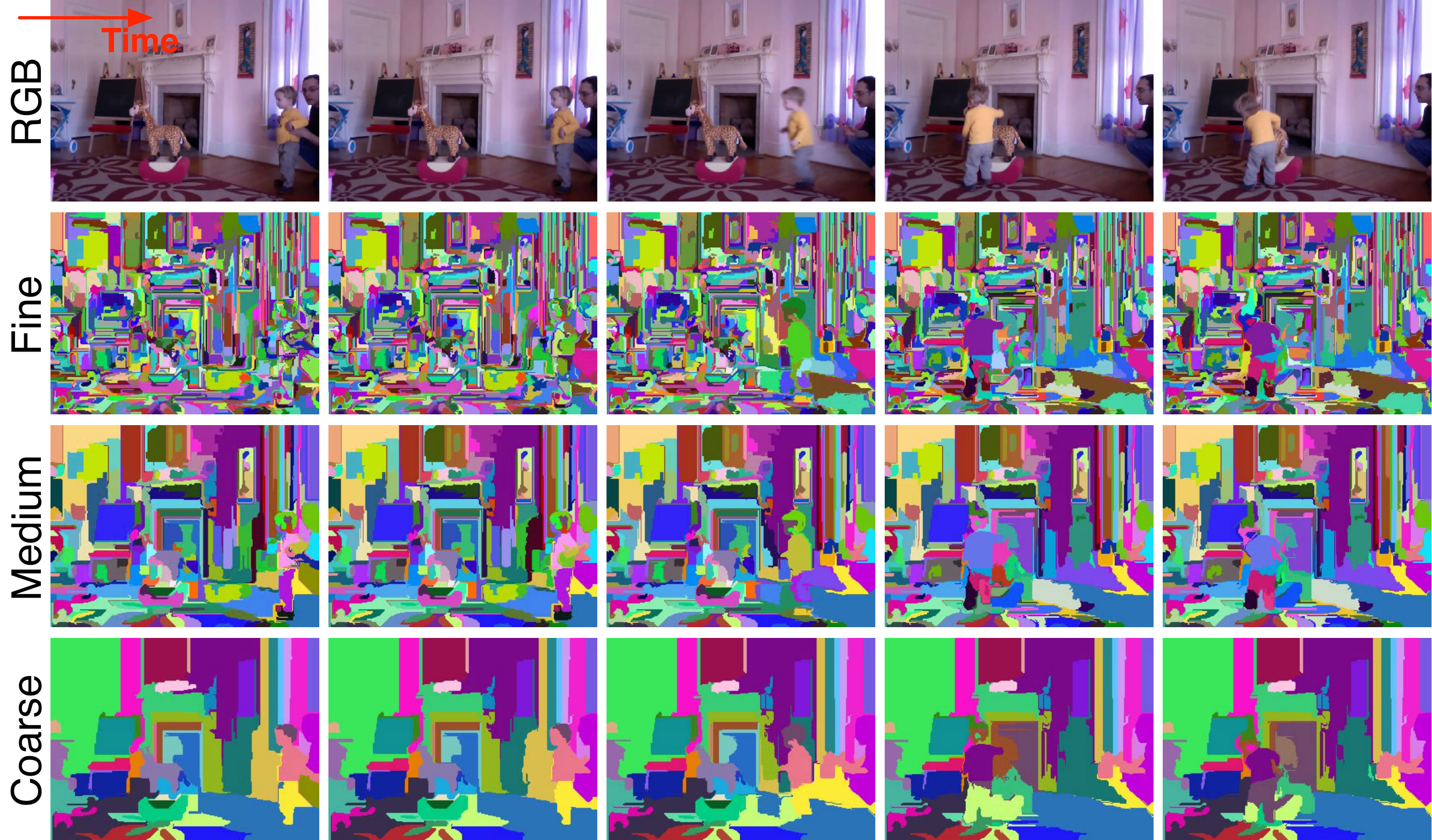}
  \caption{Example output of the streaming hierarchical supervoxel method. From left to right columns are frames uniformly sampled from a video. From top to bottom rows are: the original RGB video, the fine segmentation (low level in the hierarchy), the medium segmentation (middle level in the hierarchy), and the coarse segmentation (high level in the hierarchy).}
  \label{fig:hieseg}
\end{figure}

% Supervoxel Segmentation as Mid-level representation (motion and shape).
In contrast, segmentation of the video into spatiotemporal regions with homogeneous character, called \textit{supervoxels}, has a strong potential to overcome these limitations. Supervoxels are significantly fewer in number than the original pixels and frequently surpass the low-level features as well, and yet they capture strong features such as motion and shape, which can be used in retention of the semantics of the underlying video content. Figure \ref{fig:hieseg} shows an example supervoxel segmentation generated by the streaming hierarchical supervoxel method \cite{XuXiCoECCV2012}. The individual supervoxels are denoted as same-colored regions over time in a video. Furthermore, results in the visual psychophysics literature demonstrate that higher order processes in human perception rely on shape \cite{AmBiHaVR2012} and boundaries \cite{GrCOiN2003,NoPhRoPerception2001,OgAiCP2007}. For instance, during image/video understanding, object boundaries are interpolated to account for occlusions \cite{GrCOiN2003} and deblurred during motion \cite{OgAiCP2007}. However, the degree to which the human semantics of the video content are retained in the final segmentation is unclear. Ultimately, a better understanding of semantic retention in video supervoxel segmentation could pave the way for the future of automatic video understanding methods.
%\"{O}\v{g}men

To that end, we conduct a systematic study of how well the action and actor semantics in moving video are retained through various supervoxel segmentations. Concretely, we pose and answer the following five questions:
\newpage
\begin{enumerate}
\item Do the segmentation hierarchies retain enough information for the human 
    perceiver to discriminate actor and action?
\item How does the semantic retention vary with density of the supervoxels?
\item How does the semantic retention vary with actor?
\item How does the semantic retention vary with static versus moving background?
\item How does response time vary with action?
\end{enumerate}

% Human Study.
A preliminary version of our study appeared in \cite{XuDoHaICSC2013}, in which we present novice human observers with supervoxel segmentation videos (i.e., not RGB color videos but supervoxel segmentation videos of RGB color videos) and ask them to, as quickly as possible, determine the actor (\textit{human} or \textit{animal}) and the action (one of eight everyday actions such as \textit{walking} and \textit{eating}). The system records these human perceptions as well as the response time for them and then scores whether or not they match the ground truth perceptions; if so, then we consider that the semantics of the actor/action have been retained in the supervoxel segmentation. We systematically conduct the study with a cohort of 20 participants and 96 videos. Ultimately, the human perception results indicate that a significant amount of semantics have been retained in the supervoxel segmentation.

% Machine Study.
In addition, we conduct machine recognition experiments with a feature defined on the supervoxel segmentation, called \textit{supervoxel shape context}, and compare it with various video features, such as dense trajectories \cite{WaKlScCVPR2011} and action bank \cite{SaCoCVPR2012}. The supervoxel shape context captures the important shape information of supervoxels, which is inspired by the shape context on still images \cite{BeMaPuTPAMI2002}. Our experimental results suggest that the underlying semantics in supervoxel segmentation can be well used in machine recognition to achieve competitive results, and the overall machine recognition of actor and action seems to follow the same trend as that of human perception but more work needs to be done to get the machine recognition models up to par with the humans in terms of recognition performance.

The remainder of the paper is organized as follows. Section \ref{sec:supervoxel_segmentation} provides the background on video supervoxel segmentation. Section \ref{sec:experiment_setup} describes the details of the data set acquisition and the human perception experiment setup. Section \ref{sec:results_and_analysis} dicusses the results and our analysis of the underlying semantics in supervoxel segmentation. Section \ref{sec:machine} presents the machine recognition experiment and the results thereof. Finally, Section \ref{sec:conclusion} concludes our findings.

%-------------------------------------------------------
%-------------------------------------------------------
\section{Video Supervoxel Segmentation}
\label{sec:supervoxel_segmentation}
%-------------------------------------------------------
%-------------------------------------------------------

%-------------------------------------------------------
\subsection{Supervoxel Definition}
\label{subsec:supervoxel_definition}
%-------------------------------------------------------
Perceptual grouping of pixels into roughly homogeneous and more computationally manageable regions, called \textit{superpixels}, has become a staple of early image processing \cite{ReMaICCV2003,GoFuKoICCV2009}. Supervoxels are the video analog to the image superpixels. Recently, supervoxel segmentation has risen as a plausible first step in early video processing \cite{BrToICCV2009,GrKwHaCVPR2010,XuCoCVPR2012,XuXiCoECCV2012}. Consider the following general mathematical definition of supervoxels, as given in \cite{XuCoCVPR2012}. Given a 3D lattice $\Lambda^3$ composed by voxels (pixels in a video), a supervoxel $s$ is a subset of the lattice $s \subset \Lambda^3$ such that the union of all supervoxels comprises the lattice and they are pairwise disjoint: $\bigcup_i s_i = \Lambda^3 \wedge s_i \bigcap s_j = \varnothing \; \forall i,j \text{ pairs}$. 

Although the lattice $\Lambda^3$ itself is indeed a supervoxel segmentation, it is far from a so-called \textit{good} one \cite{XuCoCVPR2012}. Typical algorithms seek to enforce principles of spatiotemporal grouping---proximity, similarity and continuation---from classical Gestalt theory \cite{PaMIT1999,wertheimer1938laws}, boundary preservation, and parsimony. From the perspective of machine vision, the main rationale behind supervoxel oversegmentation is two fold: (1) voxels are not natural elements but merely a consequence of the discrete sampling of the digital videos and (2) the number of voxels is very high, making many sophisticated methods computationally infeasible. Therefore, supervoxels serve as an important data representation of a video, such that various image/video features may be computed on the supervoxels, including color histograms, textons, etc.

%-------------------------------------------------------
\subsection{Streaming Hierarchical Supervoxel Method}
\label{subsec:streamGBH}
%-------------------------------------------------------
We use the state of the art streaming hierarchical supervoxel method \cite{XuXiCoECCV2012} to generate a supervoxel segmentation hierarchy $\mathcal{S}=\{S^1, S^2, \dots, S^H\}$ of an input video $\mathcal{V}$, where $S^h=\bigcup_i s^h_i, h \in \{1, 2, \dots, H\}$ is the supervoxel segmentation at level $h$ in the hierarchy. The method obtains the hierarchical segmentation result $\mathcal{S}$ by minimizing:
\begin{align}
\mathcal{S}^* = \argmin_{\mathcal{S}} E(\mathcal{S}|\mathcal{V})
\enspace,
\label{eq:problem}
\end{align}
where the objective criterion $E(\cdot|\cdot)$ is defined by the minimum spanning tree method in \cite{FeHuIJCV2004}. For example, for the $h$th level in the hierarchy, the objective criterion is defined as:
\begin{align}
E(S^h|\mathcal{V}) = \tau \sum_{s \in S^h} \sum_{e \in 
\text{MST}(s)} w(e) + \sum_{s,t \in S^h} \min_{e \in <s,t>} w(e)
\enspace,
\label{eq:energy}
\end{align}
where, $\text{MST}(s)$ denotes the minimum spanning tree (of voxels or supervoxels from the previous fine level in the hierarchy) in the supervoxel $s$, $e$ is the edge defined by the 3D lattice $\Lambda^3$, $w(e)$ is the edge weight, and $\tau$ is a parameter that balances the two parts. The edge weight $w(e)$ captures the color differences of voxels. By minimizing Equation \ref{eq:problem}, the algorithm ultimately outputs a supervoxel segmentation hierarchy of the original input RGB video.

Figure \ref{fig:hieseg} shows a hierarchical supervoxel segmentation produced by \cite{XuXiCoECCV2012}. The segmentations from top to bottom rows are sampled from low, middle, and high levels in a supervoxel segmentation hierarchy, where each have fine, medium and coarse segments respectively. Each supervoxel has a unique color and we randomly color the output supervoxels in one level with the constraint that the same color is not shared by different supervoxels. In general, we allow reuse of colors in different levels in the segmentation hierarchy, since they are not used in a single run of experiment in this work. 

%-------------------------------------------------------
\subsection{Supervoxels: Rich Decompositions of RGB Videos}
\label{sec:supervoxel_hierarchy}
%-------------------------------------------------------
Considering the example in Figure \ref{fig:hieseg}, we observe that the hierarchy of the supervoxel segmentation captures different levels of semantics of the original RGB video. For example, one tends to recognize the humans easier from coarser levels in the hierarchy, since they are captured by fewer supervoxels; however, the coarser levels lose the detailed content in the video, such as the woman in the painting hanging on the wall, which is still captured at the medium level. 

\begin{figure}[!t]
  \centering
  \includegraphics[width=\textwidth]{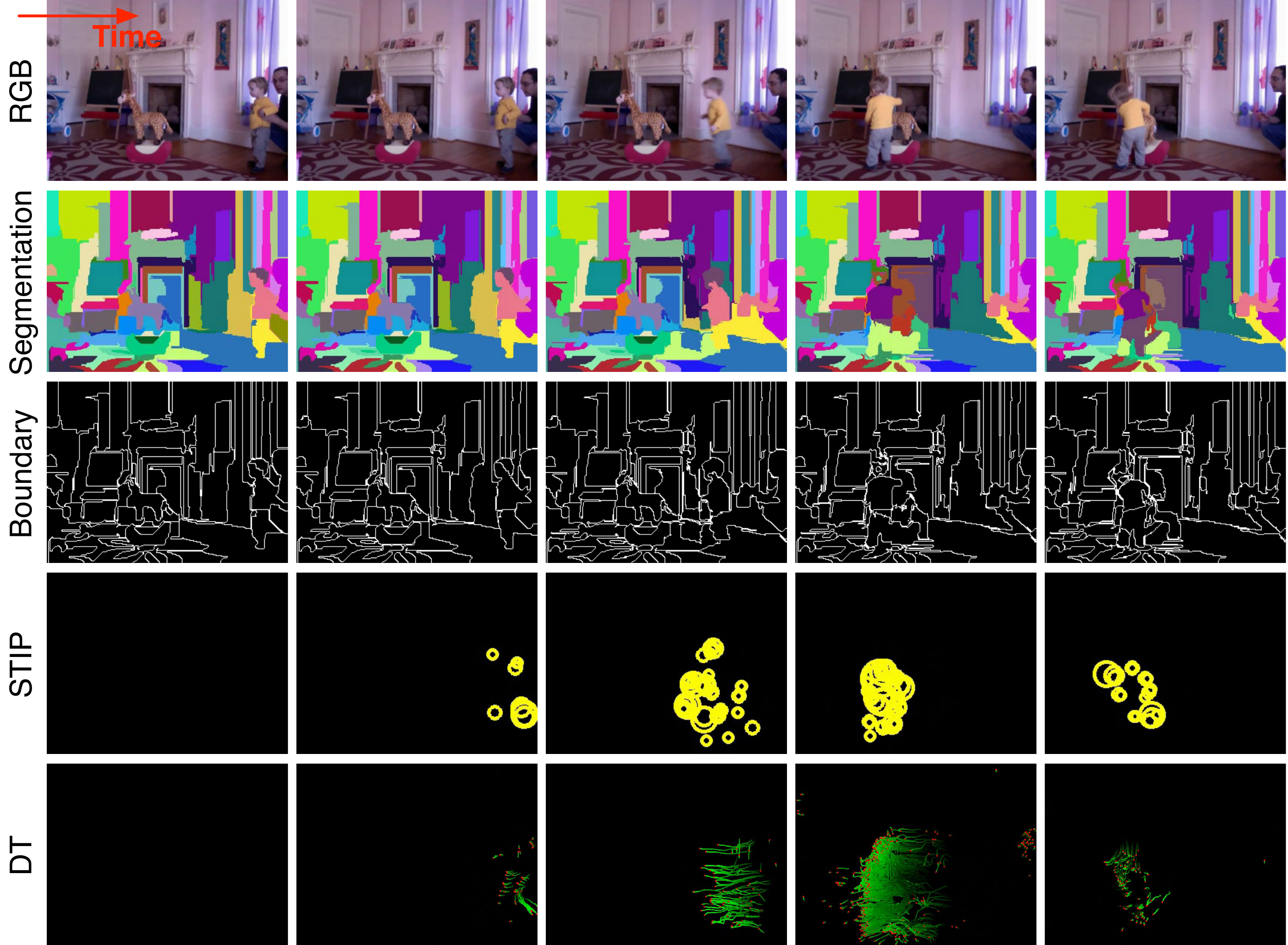}
  \caption{A comparison of different video feature representations. From top to bottom rows are: the RGB video, the supervoxel segmentation, extracted boundaries of supervoxel segmentation, space-time interest points (STIP), and dense trajectories (DT).}
  \label{fig:features}
\end{figure}

Compared with other features, such as space-time interest points (STIP) \cite{LaIJCV2005} and dense trajectories (DT) \cite{WaKlScCVPR2011}, which are frequently used in video analysis \cite{TaAlYuCVPR2012}, the supervoxel segmentation seems to retain more semantics of the RGB video (in this paper we seek to quantify how many of these semantics are retained for one set of actors and actions). Figure \ref{fig:features} shows a visual comparison among those features. STIP and DT use the sampled points and trajectories as the data representation---this is not the full STIP or DT feature descriptor representation, which also measures other information, such as gradient. We will detail it in Section \ref{sec:machine}.

By only watching the videos of STIP and DT, as shown in the bottom two rows of Figure \ref{fig:features}, it seems unlikely that humans could recover the content of a video, especially when there is little motion in a video. On the other hand, one can easily recover the content of a video by watching the supervoxel segmentation video, likely due to the fact that the supervoxel segmentation retains the shape of the objects (boundaries of the supervoxel segmentation are also shown in the third row of Figure \ref{fig:features}). Zitnick and Parikh \cite{ZiPaCVPR2012} show that the segmentation boundaries are in general better than classical edge detection methods, such as those generated by the Canny edge detector\cite{CaTPAMI1986}, for automatic image understanding, and they perform as well as humans using low-level cues. The precise goal of this paper is to explore exactly how much semantic content, specifically the actor (human or animal) and the action, is retained in the supervoxel segmentation. We describe the experiment and results of supervoxel human perception in the next two sections.

%-------------------------------------------------------
%-------------------------------------------------------
\section{Experiment Setup}
\label{sec:experiment_setup}
%-------------------------------------------------------
%-------------------------------------------------------
We have set up a systematic experiment to study actor and action semantic retention in the supervoxel segmentation. By actor we simply mean \textit{human} or \textit{animal}. For action, we include a set of eight actions: \textit{climbing}, \textit{crawling}, \textit{eating}, \textit{flying}, \textit{jumping}, \textit{running}, \textit{spinning} and \textit{walking}. We have gathered a complete set of videos (Section \ref{subsec:data_set}) and processed them through the segmentation algorithm (Section \ref{sec:supervoxel_segmentation}). Then we show the  segmentation videos to human observers (Section \ref{subsec:study_cohort}) and request them to make a forced-choice selection of actor and action (Section \ref{subsec:collection_site}). Finally, we analyze the aggregate results over the full data cohort and quantify the retention of semantics (Section \ref{sec:results_and_analysis}).

%-------------------------------------------------------
\subsection{Data Set}
\label{subsec:data_set}
%-------------------------------------------------------

\begin{figure}[!t]
  \centering
  \includegraphics[width=\textwidth]{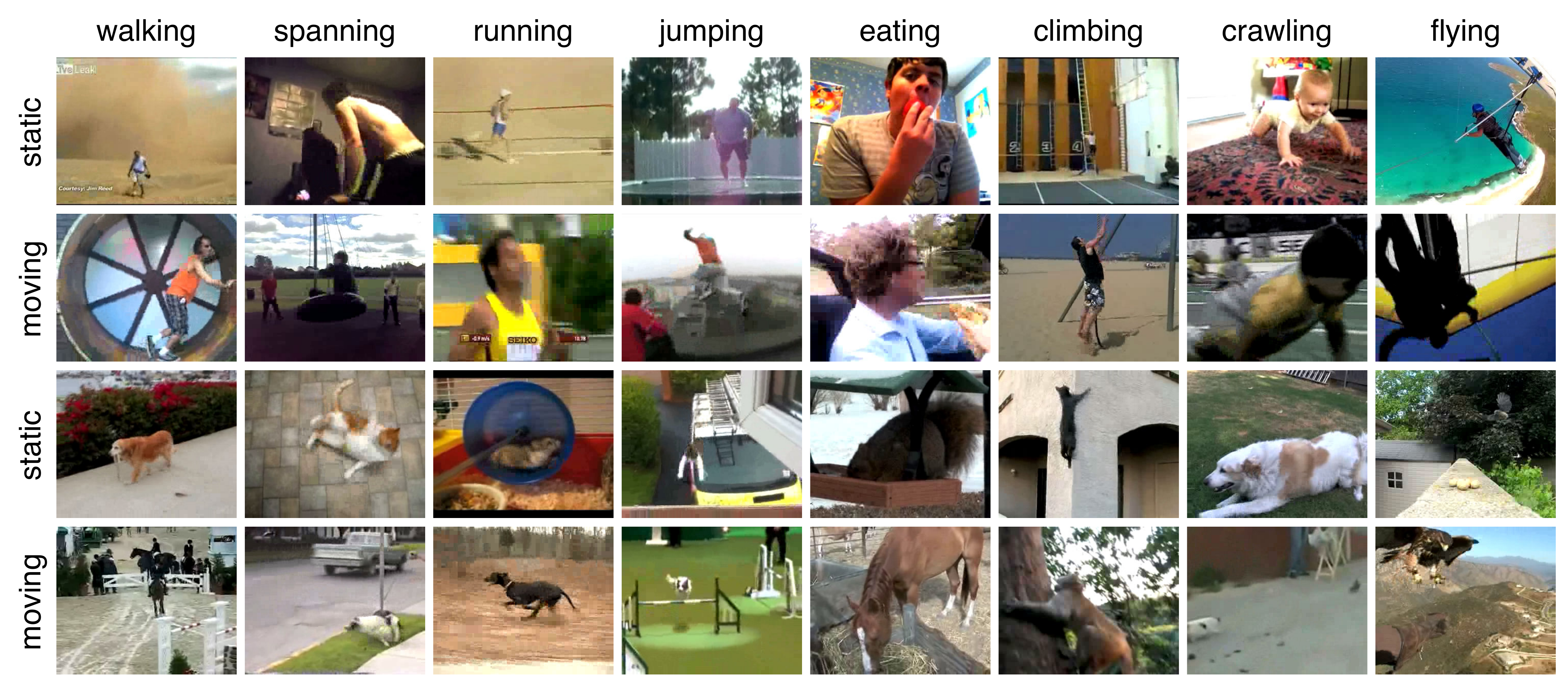}
  \caption{A snapshot of the RGB videos in our data set. The actors in the top two rows are humans and in the bottom two rows are animals. The data set consists of two kinds of actors, eight actions and two types of background settings, resulting in a total of 32 videos.}
  \label{fig:dataset}
\end{figure}

%-------------------------------------------------------
\subsubsection{Data Collection}
\label{subsubsect:data_collection}
%-------------------------------------------------------
We have collected a data set with two kinds of actors (\textit{humans} and \textit{animals}) performing eight different actions: \textit{climbing}, \textit{crawling}, \textit{eating}, \textit{flying}, \textit{jumping}, \textit{running}, \textit{spinning} and \textit{walking}. We only include animals that frequently appear in (North American) daily life, such as dogs, cats, birds, squirrels and horses. The backgrounds of the videos fall into two categories: static (relatively static objects such as ground and buildings with little camera changes) and moving (moving objects in the background, such as in a traffic or dramatic camera changes). A complete RGB video data set consists of 32 videos in total (2 actors $\times$ 8 actions $\times$ 2 background types $= 32$). Figure \ref{fig:dataset} shows a snapshot of the RGB videos we collected. 

Each video in the data set is about four seconds long and the actor starts the action immediately after the video plays. We, however, show the videos at half-frame-rate when conducting the experiment to allow ample response time for the human participants. We have attempted to exclude those videos that have ambiguity with either the actors or the actions, and only use videos that have a major actor performing one single action. For example, a disqualified human jumping video usually contains the running before jumping. But, some ambiguity remains due to the general complexity of dynamic video.  The data set used in this paper is a complete data set having a single video for each actor, action and background type tuple. All videos were downloaded from public ``wild'' repositories, such as YouTube. %\footnote{Contact the authors if you want a copy of the data set.}

For each of the RGB videos, we use the method described in Section \ref{subsec:streamGBH} to obtain a supervoxel segmentation hierarchy. We first use ffmpeg to resize the videos to 320x240 maintaining the original aspect ratio. Then, we run the \textit{gbh\_stream} program (LIBSVX version 2.0\footnote{\url{http://www.cse.buffalo.edu/~jcorso/r/supervoxels/}}) with the following parameters: \texttt{c: 0.2}, \texttt{c\_reg: 10}, \texttt{min: 20}, \texttt{sigma: 0.4}, \texttt{range: 10}, \texttt{hie\_num: 30}.  Note that \texttt{c} and \texttt{c\_reg} map to $\tau$ in Eq. \ref{eq:energy} (\texttt{c} is used at the first level and \texttt{c\_reg} is used at all other hierarchy levels).

We sample three different levels (fine: 8th level, medium: 16th level, coarse: 24th level) from the hierarchy, similar as in Figure \ref{fig:hieseg}. Therefore, the full set of data we used to run the semantic retention experiment is the 96 supervoxel segmentation videos.  Note that the audio is disabled, so that the participants only have the vision perception of the supervoxel segmentation videos (and never the RGB videos).

%-------------------------------------------------------
\subsubsection{Data Split}
\label{subsubsec:data_split}
%-------------------------------------------------------
We create a threeway split of the 96 videos into three sets: alpha, beta and gamma. Since each of the original 32 videos is represented in three levels of the hierarchy, it is imperative to make the threeway split and thereby avoid one participant seeing the same video twice but on two different supervoxel hierarchy levels. So, each of alpha, beta and gamma have the full 32 videos, but on different hierarchy levels (and uniformly varying over levels in each of the three splits). Based on the ordering in the database, alpha will start with one level (say coarse) in the hierarchy, then beta will have the next (medium) and gamma the third (fine) for one original RGB video. For the next original RGB video, it will rotate, so that alpha has the next level (now medium), beta the next (fine) and gamma will wrap around to the first (coarse) again. This repeats through all 96 supervoxel videos. Before the videos are shown to the participant, the order of the videos is shuffled, so that the participant cannot deduce the contents based on an ordering of the videos (like human human human ... animal animal animal).

%-------------------------------------------------------
\subsection{Study Cohort}
\label{subsec:study_cohort}
%-------------------------------------------------------
The study cohort is 20 college-age participants. To ensure generality, we exclude those students who are studying video segmentation (and hence may have already developed an \textit{eye} for semantic content in supervoxel segmentations). Each participant is shown one split of the videos (alpha, beta or gamma).  And each participant sees a given video only once. Participants never see the input RGB videos.

%-------------------------------------------------------
\subsection{Human User Interface and Instructions}
\label{subsec:collection_site}
%-------------------------------------------------------
\begin{figure}[!t]
  \centering
  \includegraphics[width=\textwidth]{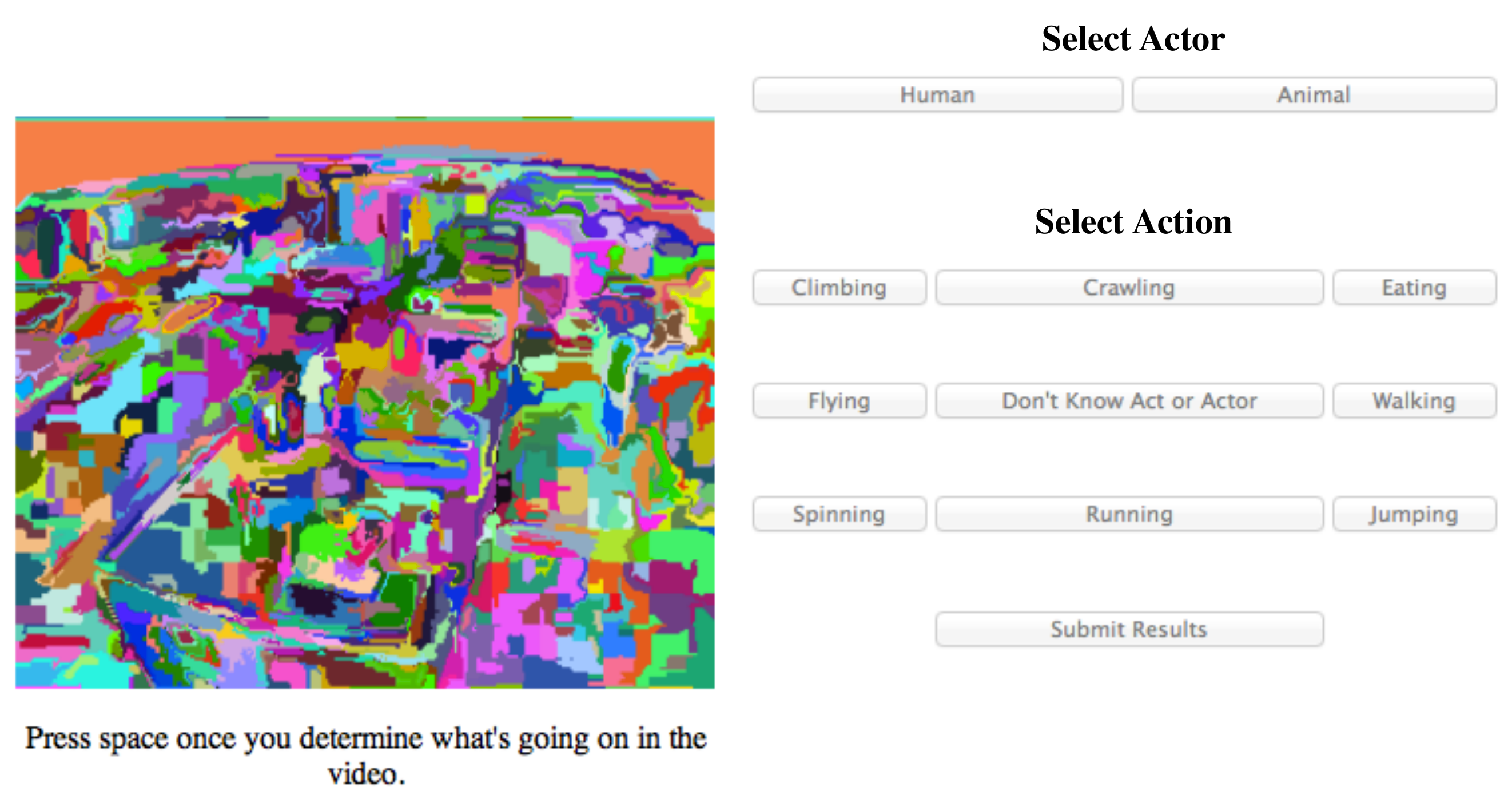}
  \caption{A snapshot of the user interface for the human perception experiment.}
  \label{fig:site}
\end{figure}

The user interface is web-based. Figure \ref{fig:site} shows a snapshot of it. The left part of the participant's screen is the supervoxel segmentation video and the right part of the participant's screen comprises two sets of buttons that allow the user to choose either human or animal as the actor, and to choose one of the eight actions. The participant has the option to choose \textit{unknown} (the option ``Don't know act or actor'' is shown in the center of the select action area in Figure \ref{fig:site}). Such an \textit{unknown} option prevents the participant from random selection.

Initially, when the participant starts the experiment, the left part of the screen is blank and buttons on the right side are deactivated (grayed out); once the next video in their split is downloaded locally, it prompts the user with a ``ready'' message. As soon as the participant presses the \textit{space} key, it starts to show the supervoxel segmentation video and the interface triggers a timer that records the response time of the participant. The participant is required to respond by pressing the \textit{space} key again as soon as he or she captured enough information to reach a decision (i.e., knows the actor and action in the supervoxel video). The amount of time between these two \textit{space} key pressing is recorded as one's response time. After this second \textit{space} key is hit, the buttons on the right side are activated and ready for the participant to select them.   The participant can only watch the video once, which means once the video reaches the end, the participant is forced to make a decision (or choose unknown) without the option to watch it again. In this case, the whole video time is recorded as one's response time.  This process is repeated for each video in the split (alpha, beta or gamma) until the end.

Before a participant begins, s/he is instructed briefly as to the nature of the experiment (trying to recognize the actor and action in a supervoxel video) and walked through the user interface. They are instructed that time is recorded and important; they should hence stop the video as soon as they know the answer. They are not shown any example supervoxel video before the experiment starts.

%-------------------------------------------------------
%-------------------------------------------------------
\section{Results and Analysis}
\label{sec:results_and_analysis}
%-------------------------------------------------------
%-------------------------------------------------------
The response of a single video by one participant is defined as a \textit{supervoxel human perception}: $<\textit{actor}, \textit{action}, \textit{response time}>$. In total, we have 640 supervoxel human perceptions collected (32 videos in each split $\times$ 20 participants). The original RGB videos are used as the ground truth data to measure the \textit{match} of the supervoxel human perceptions. We also measure the response time of the participants for both \textit{matched} perceptions and \textit{unmatched} perceptions. Our analysis is organized systematically according to five key questions regarding semantic retention.

%-------------------------------------------------------
\subsection{Do the segmentation hierarchies retain enough information for the 
human perceiver to discriminate actor and action?}
%-------------------------------------------------------

% Actor Discrimination
\begin{table}
\caption{Confusion matrix for actor discrimination.}
\label{table:actor_discrimination}
\centering
{
\small
\begin{tikzpicture}
  \tikzset{square matrix/.style={
    matrix of nodes,
    column sep=-\pgflinewidth, row sep=-\pgflinewidth,
    nodes={draw,minimum height=#1,anchor=center,text width=#1,align=center,inner 
    sep=0pt},
    font=\sf,
  },
  square matrix/.default=2.5em
  }

\matrix[square matrix] (m)
{
|[fill=white ]| 0  & |[fill=white ]| 0  & |[fill=white ]| 0  \\
|[fill=gray!27 ]| 0.11  & |[fill=gray!219 ,text=white]| 0.86  & |[fill=gray!7 ]| 
0.03  \\
|[fill=gray!42 ]| 0.17  & |[fill=gray!12 ]| 0.05  & |[fill=gray!200 
,text=white]| 0.78  \\
};

\node[left=1pt of m-1-1, font=\sf] {unknown};
\node[left=1pt of m-2-1, font=\sf] {human};
\node[left=1pt of m-3-1, font=\sf] {animal};
\node[above=1pt of m-1-1, font=\sf] {un};
\node[above=1pt of m-1-2, font=\sf] {hu};
\node[above=1pt of m-1-3, font=\sf] {an};

\end{tikzpicture}
}
\end{table}

%-------------------------------------------------------
\subsubsection{Actor discrimination}
\label{subsubsec:human_actor}
%-------------------------------------------------------
Table \ref{table:actor_discrimination} shows a confusion matrix of the actor discrimination. As high as 86\% of the human perceptions correctly identify the human actors, 78\% for the animal actors, and in average 82\% for actors in general. We also note that participants tend to choose the \textit{unknown} option when they are less confident of the supervoxel segmentation. There is only a small portion of unmatched perceptions 3\% and 5\% that mistake human as animal or vice versa. This is hence strong evidence showing that the supervoxel segmentation indeed has the ability to retain the actor semantics from the original RGB videos. We suspect this binary discrimination performance is so high because the data cohort includes videos with one dominant actor and the human participant is able to localize this actor with the supervoxel motion information and then use the supervoxel shape information to determine human or animal. We suspect the reason why the human perception of animal actors is less than that of human actors is because the animals in the data set vary more broadly in relative location and orientation than the humans do.

%-------------------------------------------------------
\subsubsection{Action discrimination}
\label{subsubsec:human_action}
%-------------------------------------------------------
Table \ref{table:action_discrimination} shows a confusion matrix of the action discrimination. The top three scoring actions are climbing (90\%), running (79\%), and eating (76\%), while the bottom three ones are walking (57\%), jumping (57\%), and spinning (65\%). On average, 70.4\% of human perceptions correctly match the actions. Of the lower performing actions, only walking has been easily confused with the other actions (12\% to spinning and 12\% to running, which may be due to semantic ambiguity---see the example of the human walking in the spinning wheel in Figure \ref{fig:visual}); jumping and spinning have more \textit{unknowns} (19\% and 15\% respectively) rather than being confused with other actions. An interesting point to observe is that running and climbing are perceived \textit{unknown} significantly fewer than the other six actions. We suspect this is due to the dominant unidirectional motion of these two actions. Overall, this is more strong evidence that suggests the supervoxel segmentation can well retain the action semantics from the original RGB videos.

% Action Discrimination
\begin{table}
\caption{Confusion matrix for action discrimination.}
\label{table:action_discrimination}
\centering
{
\small
    \begin{tikzpicture}
    \tikzset{square matrix/.style={
        matrix of nodes,
        column sep=-\pgflinewidth, row sep=-\pgflinewidth,
        nodes={draw,
          minimum height=#1,
          anchor=center,
          text width=#1,
          align=center,
          inner sep=0pt
        },
        font=\sf,
      },
      square matrix/.default=2.5em
    }

    \matrix[square matrix] (m)
    {
|[fill=white ]| 0  & |[fill=white ]| 0  & |[fill=white ]| 0  & |[fill=white ]| 0  
& |[fill=white ]| 0  & |[fill=white ]| 0  & |[fill=white ]| 0  & |[fill=white ]| 
0  & |[fill=white ]| 0  \\
|[fill=gray!28 ]| 0.11  & |[fill=gray!146 ,text=white]| 0.57  & |[fill=gray!31 
]| 0.12  & |[fill=gray!31 ]| 0.12  & |[fill=white ]| 0  & |[fill=gray!3 ]| 0.01  
& |[fill=gray!3 ]| 0.01  & |[fill=gray!9 ]| 0.04  & |[fill=white ]| 0  \\
|[fill=gray!38 ]| 0.15  & |[fill=gray!15 ]| 0.06  & |[fill=gray!165 
,text=white]| 0.65  & |[fill=gray!6 ]| 0.03  & |[fill=white ]| 0  & |[fill=white 
]| 0  & |[fill=gray!3 ]| 0.01  & |[fill=gray!9 ]| 0.04  & |[fill=gray!15 ]| 0.06  
\\
|[fill=gray!3 ]| 0.01  & |[fill=gray!19 ]| 0.07  & |[fill=gray!19 ]| 0.07  & 
|[fill=gray!200 ,text=white]| 0.79  & |[fill=gray!9 ]| 0.04  & |[fill=white ]| 0  
& |[fill=white ]| 0  & |[fill=gray!3 ]| 0.01  & |[fill=white ]| 0  \\
|[fill=gray!47 ]| 0.19  & |[fill=gray!3 ]| 0.01  & |[fill=gray!9 ]| 0.04  & 
|[fill=gray!22 ]| 0.09  & |[fill=gray!146 ,text=white]| 0.57  & |[fill=white ]| 
0  & |[fill=white ]| 0  & |[fill=gray!3 ]| 0.01  & |[fill=gray!22 ]| 0.09  \\
|[fill=gray!47 ]| 0.19  & |[fill=white ]| 0  & |[fill=white ]| 0  & |[fill=white 
]| 0  & |[fill=white ]| 0  & |[fill=gray!194 ,text=white]| 0.76  & |[fill=gray!9 
]| 0.04  & |[fill=white ]| 0  & |[fill=gray!3 ]| 0.01  \\
|[fill=gray!15 ]| 0.06  & |[fill=gray!3 ]| 0.01  & |[fill=white ]| 0  & 
|[fill=white ]| 0  & |[fill=gray!6 ]| 0.03  & |[fill=white ]| 0  & 
|[fill=gray!229 ,text=white]| 0.90  & |[fill=white ]| 0  & |[fill=white ]| 0  \\
|[fill=gray!51 ]| 0.20  & |[fill=gray!6 ]| 0.03  & |[fill=white ]| 0  & 
|[fill=gray!15 ]| 0.06  & |[fill=gray!3 ]| 0.01  & |[fill=white ]| 0  & 
|[fill=gray!3 ]| 0.01  & |[fill=gray!175 ,text=white]| 0.69  & |[fill=white ]| 0  
\\
|[fill=gray!47 ]| 0.19  & |[fill=gray!6 ]| 0.03  & |[fill=gray!3 ]| 0.01  & 
|[fill=white ]| 0  & |[fill=gray!3 ]| 0.01  & |[fill=gray!3 ]| 0.01  & 
|[fill=gray!6 ]| 0.03  & |[fill=gray!6 ]| 0.03  & |[fill=gray!178 ,text=white]| 
0.70  \\
};

\node[left=1pt of m-1-1, font=\sf] {unknown};
\node[left=1pt of m-2-1, font=\sf] {walking};
\node[left=1pt of m-3-1, font=\sf] {spinning};
\node[left=1pt of m-4-1, font=\sf] {running};
\node[left=1pt of m-5-1, font=\sf] {jumping};
\node[left=1pt of m-6-1, font=\sf] {eating};
\node[left=1pt of m-7-1, font=\sf] {climbing};
\node[left=1pt of m-8-1, font=\sf] {crawling};
\node[left=1pt of m-9-1, font=\sf] {flying};
\node[above=1pt of m-1-1, font=\sf] {un};
\node[above=1pt of m-1-2, font=\sf] {wl};
\node[above=1pt of m-1-3, font=\sf] {sp};
\node[above=1pt of m-1-4, font=\sf] {rn};
\node[above=1pt of m-1-5, font=\sf] {jm};
\node[above=1pt of m-1-6, font=\sf] {ea};
\node[above=1pt of m-1-7, font=\sf] {cl};
\node[above=1pt of m-1-8, font=\sf] {cr};
\node[above=1pt of m-1-9, font=\sf] {fl};

\end{tikzpicture}
}
\end{table}

%-------------------------------------------------------
\subsection{How does the semantic retention vary with density of the supervoxels?}
%-------------------------------------------------------

\begin{figure}[!t]
  \centering
  \includegraphics[width=0.75\textwidth]{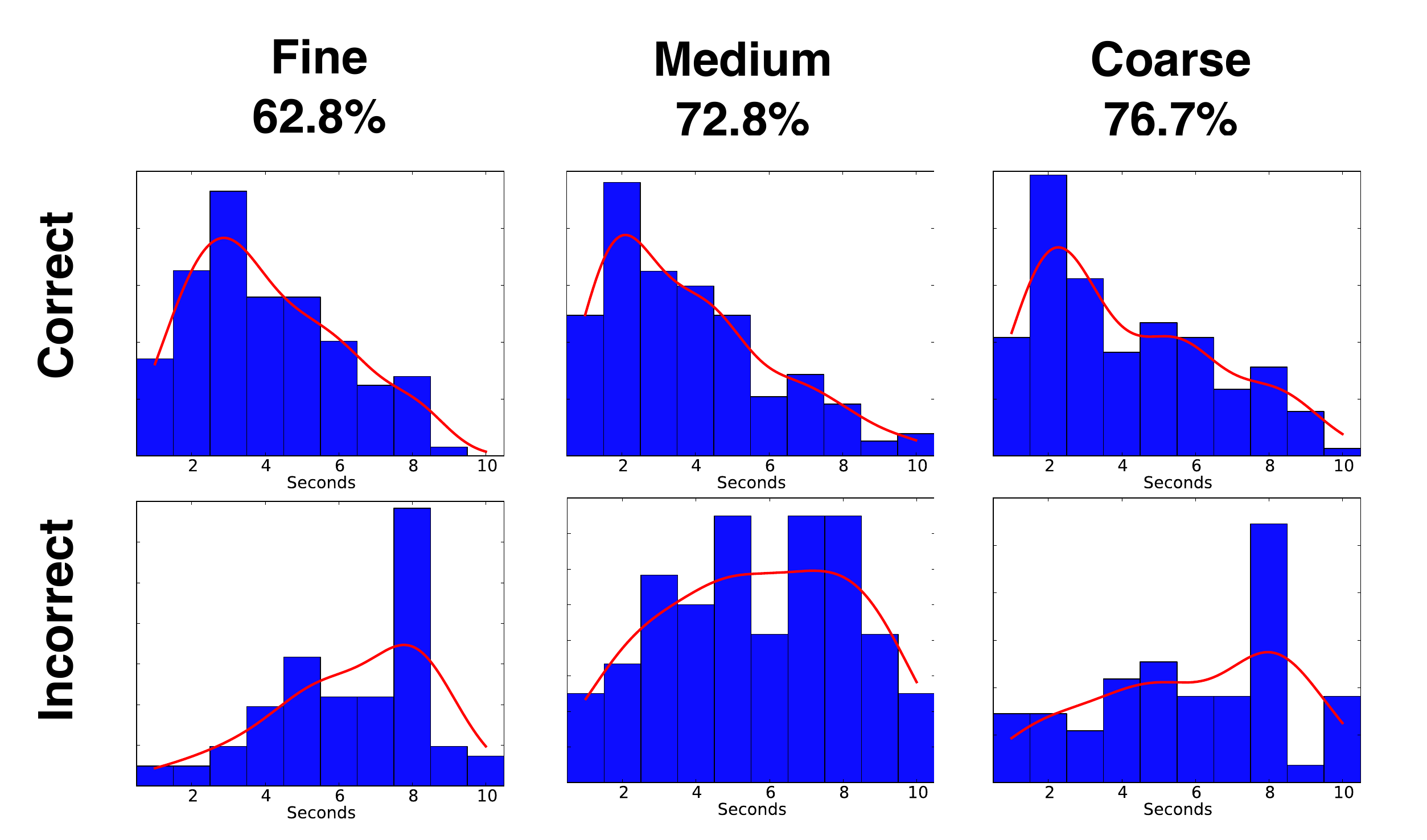}
  \caption{The performance of supervoxel semantic retention of actor and action on three levels from the supervoxel segmentation hierarchy: fine, medium and coarse. The percentages on top are computed when the supervoxel human perceptions are correctly matched to ground truth ones. The middle and bottom rows are the response time figures when the supervoxel human perceptions are correctly matched and incorrectly matched respectively.}
  \label{fig:result_Level}
\end{figure}

Following the discussion of supervoxel hierarchy in Section \ref{sec:supervoxel_hierarchy}, we seek to understand how the supervoxel size influences retention of action and actor semantics. Recall that we sampled three levels from the supervoxel hierarchy to obtain fine, medium and coarse level supervoxel segmentations. Figure \ref{fig:result_Level} shows the overall performance of the supervoxel human perceptions on different levels. The percentage of correctly matched human perceptions increases when the size of supervoxels grows, suggesting that coarse segmentations more readily retain the semantics of the action and that even coarser segmentations could perform better (i.e., the perfect segmentation of the actor performing the action would likely perform best). A second plausible explanation is that for actor and action discrimination the finer details in the other levels are unlikely to be needed.

We also show study of the response time in Figure \ref{fig:result_Level}. Here, we plot the density of responses (horizontal axis is time, at half-frame-rate; vertical axis is density).  The blue bars are a simple histogram and the red curve is a Gaussian kernel density estimate. For correct action matches, the response distributions are nearly equivalent, and are heavily weighted toward the shorter end of the plot, indicating that if the participant knows the answer then typically knows it quickly. However, for the incorrect matches, we see different patterns, the fine videos are peaked at about eight seconds, which is the maximum length for most videos, indicating the participant watched the whole video and still got the wrong action perception. For fine videos, one expects this due to the great number of supervoxels being perceived, which introduces more noise. The medium and coarse scales are more uniformly distributed (although the coarse scale also has a peak at eight seconds), indicating that sometimes the perception was simply wrong. This may either be due to intrinsic limitation of the supervoxels to retain some action semantics or due to the ambiguities of the specific videos in the data set, which, although we did try to avoid, are present in some few cases. Further study on this point is needed to better understand the source of the error.

%-------------------------------------------------------
\subsection{How does the semantic retention vary with actor?}
\begin{figure}[!t]
  \centering
  \includegraphics[width=0.75\textwidth]{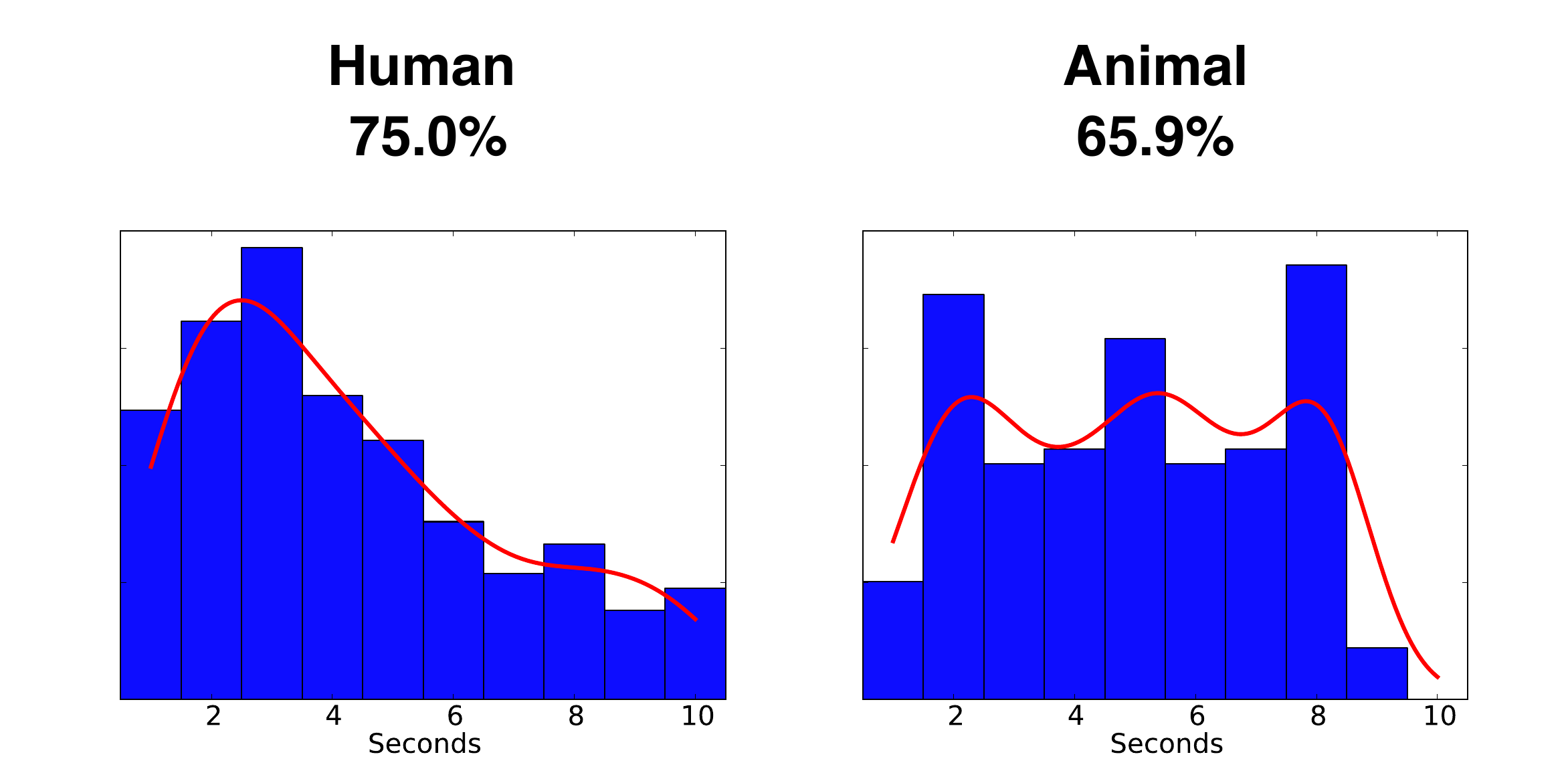}
  \caption{Performance comparison between human actors and animal actors. The percentages on top are computed when the supervoxel human perceptions are correctly matched to ground truth ones. The response time plots include both correctly and incorrectly matched supervoxel human perceptions.}
  \label{fig:result_Actor}
\end{figure}

%-------------------------------------------------------
We stratify the accuracy of the matches according to the actor performing the action.  Figure \ref{fig:result_Actor} shows the overall performance by human actors and animal actors. In general, human perception of human actors has higher match than that of animal actors. For speed, the response time of human actors has only one peak at around three seconds, while that of animal actors has multiple peaks, which may be due to the greater variation in appearance of animals in the data set than of humans. Moreover, human activity is easier to perceive than animal as studied by Pinto and Shiffrar \cite{PiShSN2009}. Considering the result in Table \ref{table:actor_discrimination}, the result in Figure \ref{fig:result_Actor} also suggests a correlation between knowing the actor and recognizing the action correctly.

%-------------------------------------------------------
\subsection{How does the semantic retention vary with static versus moving background?}
\begin{figure}[!t]
  \centering
  \includegraphics[width=0.75\textwidth]{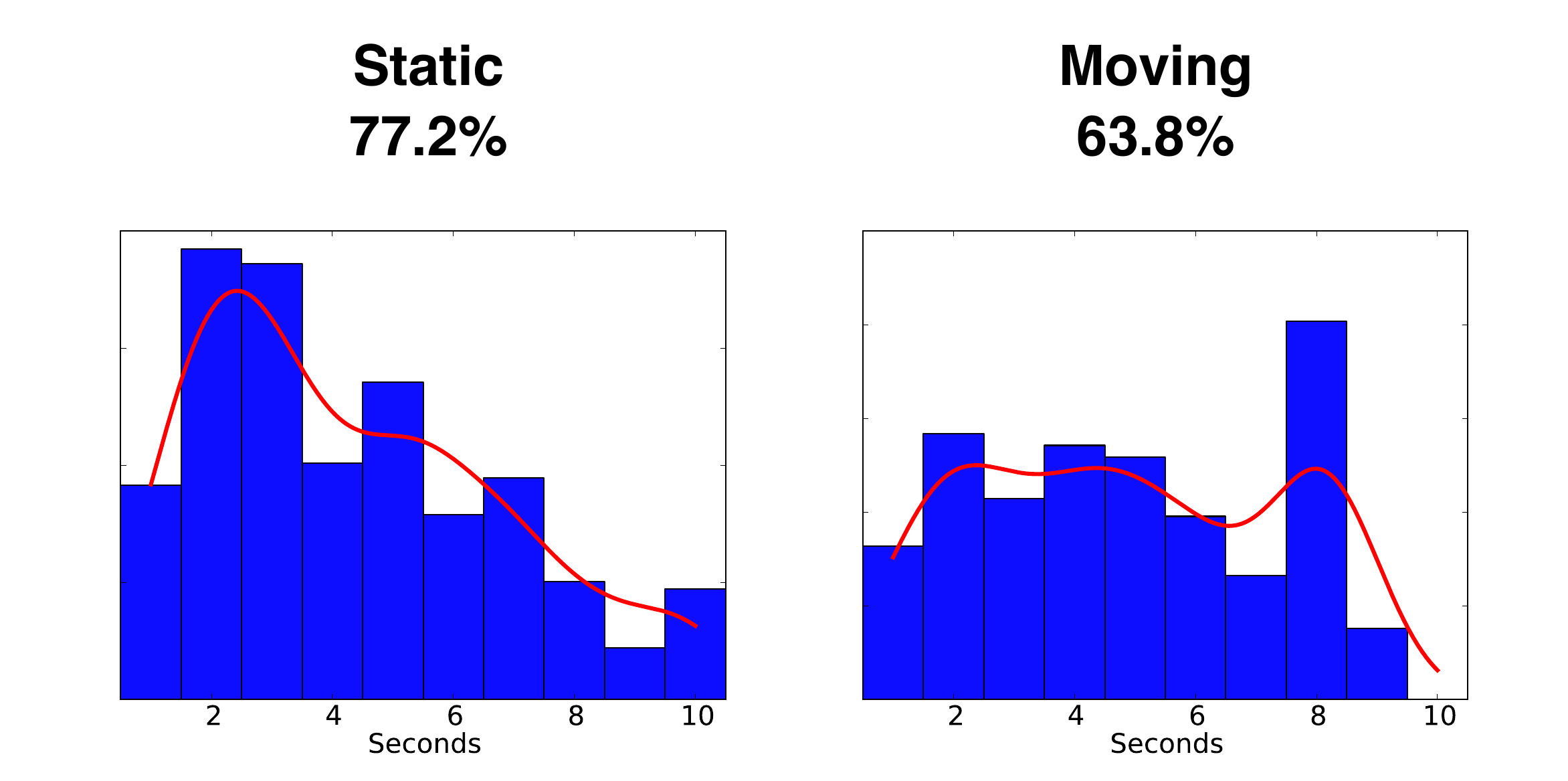}
  \caption{Performance comparison between static background and moving background. The percentages on top are computed when the supervoxel human perceptions are correctly matched to ground truth ones. The response time plots include both correctly and incorrectly matched supervoxel human perceptions.}
  \label{fig:result_Background}
\end{figure}

%-------------------------------------------------------
Figure \ref{fig:result_Background} shows the overall performance of the supervoxel human perception match for static background and moving background. Supervoxel human perceptions have higher match and shorter response time in the case of static background, as expected (the dominant actor is more easily picked out by the participant). The relatively ``flat'' curve in moving background suggests the response time for a single video highly depends on the specific background within that video.

%-------------------------------------------------------
\subsection{How does response time vary with action?}
%-------------------------------------------------------

% Time by Action
\begin{figure}[t!]
  \centering
  \includegraphics[width=\textwidth]{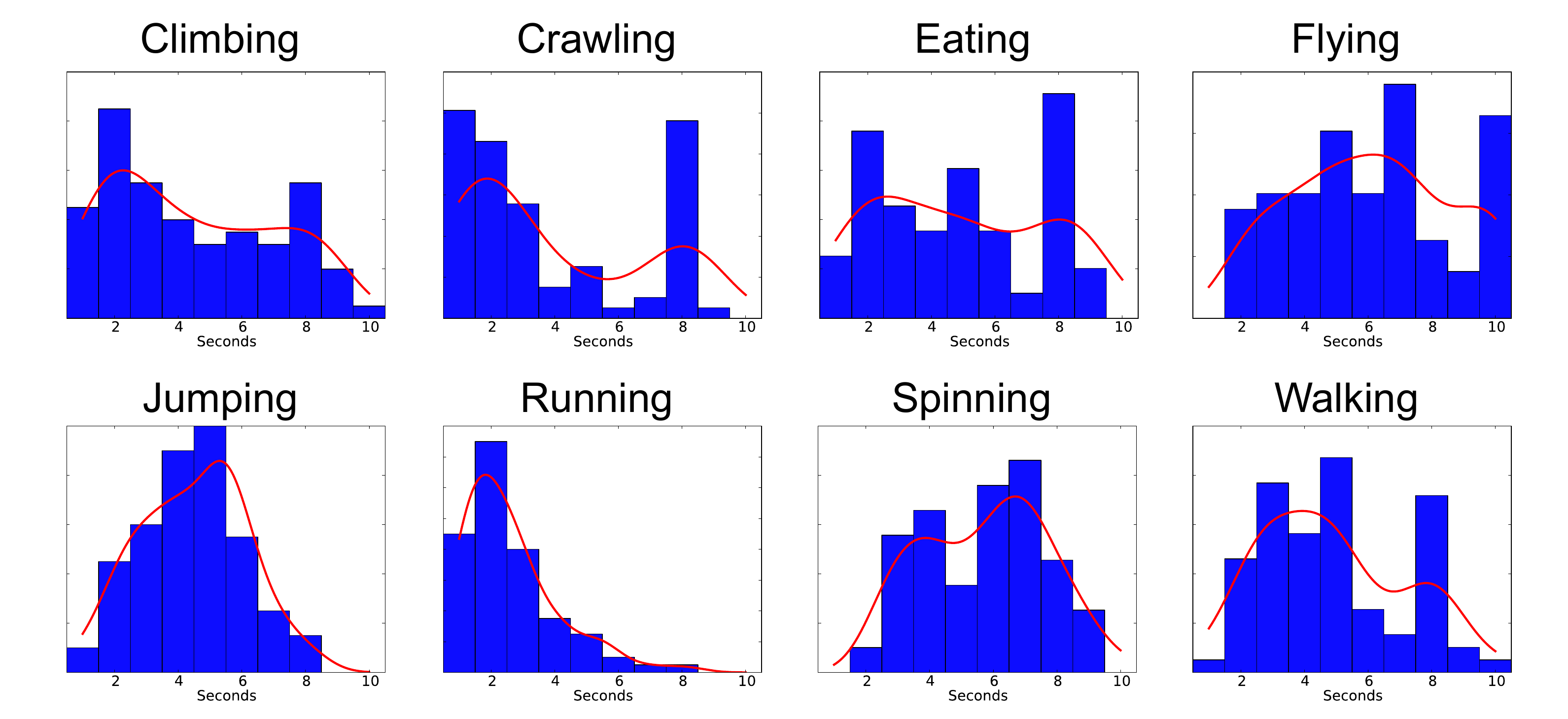}
  \caption{Response time of eight different actions for both correctly and incorrectly matched supervoxel human perceptions.}
  \label{fig:result_TimeByAction}
\end{figure}

\begin{figure}[t!]
  \begin{center}
    \small
    \begin{tabular}{cc}
    \includegraphics[width=0.485\linewidth]{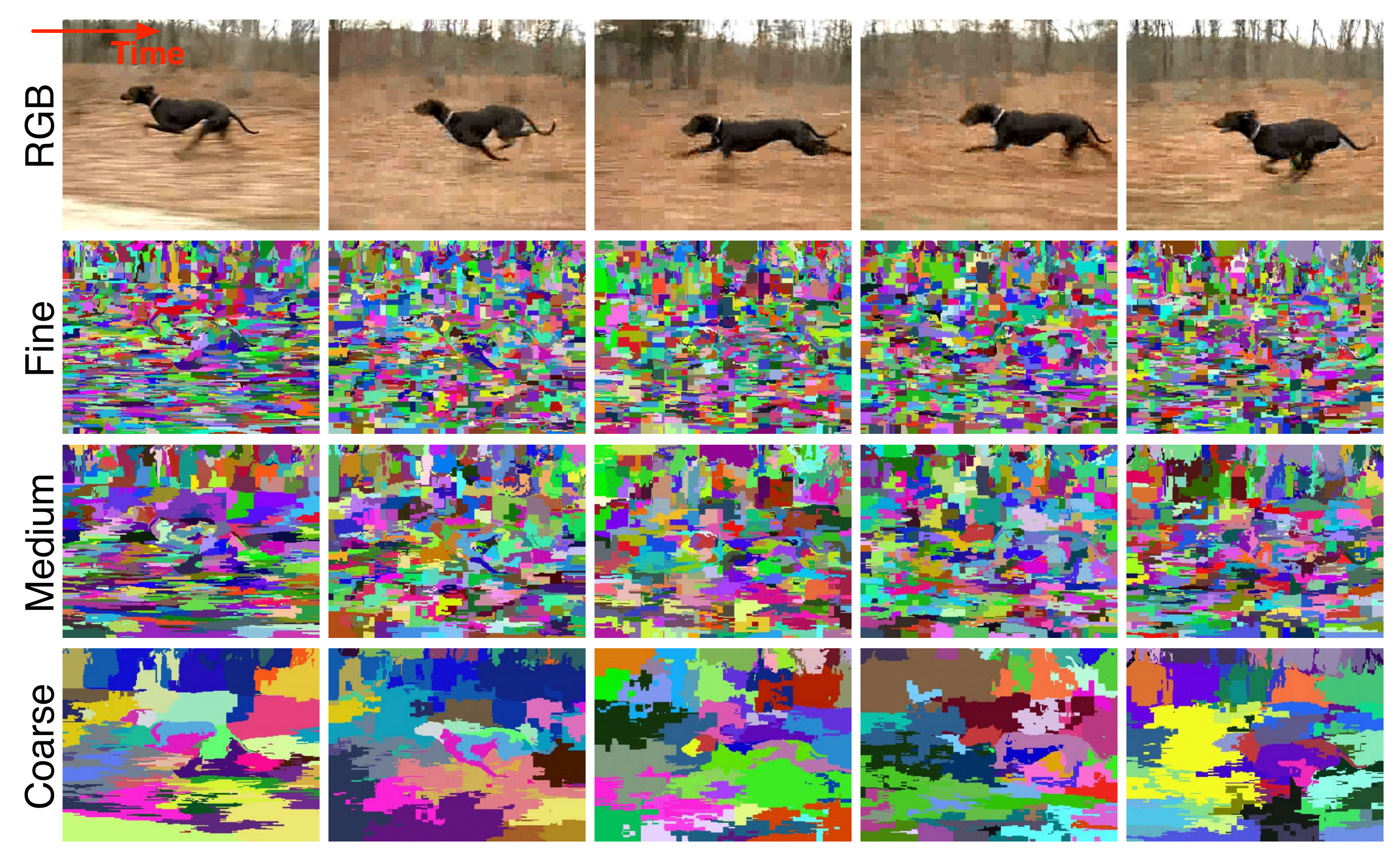} &
    \includegraphics[width=0.485\linewidth]{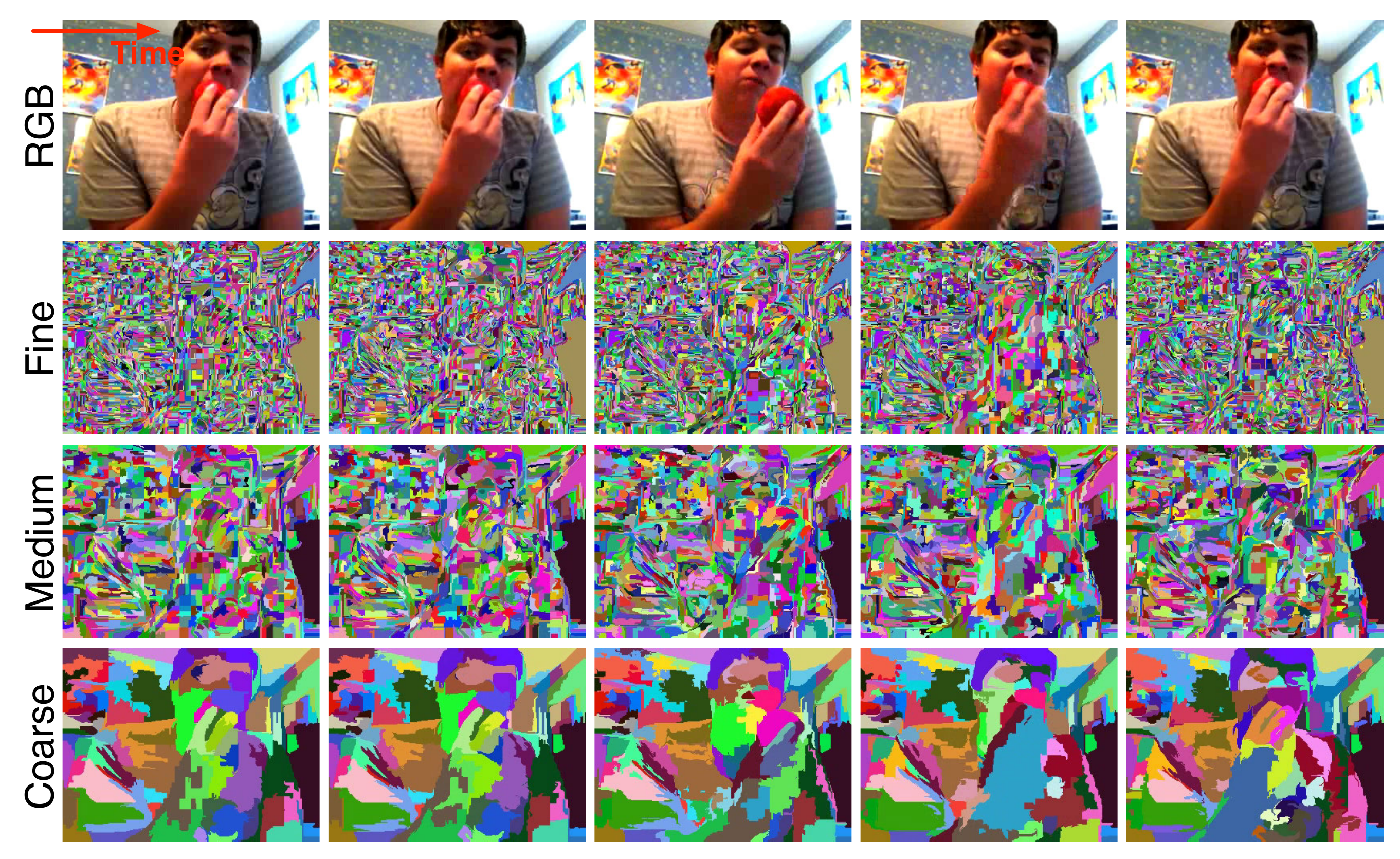} \\
    Animal\_Running\_Moving: 95\% Correct & Human\_Eating\_Static: 100\% Correct \\
    \includegraphics[width=0.485\linewidth]{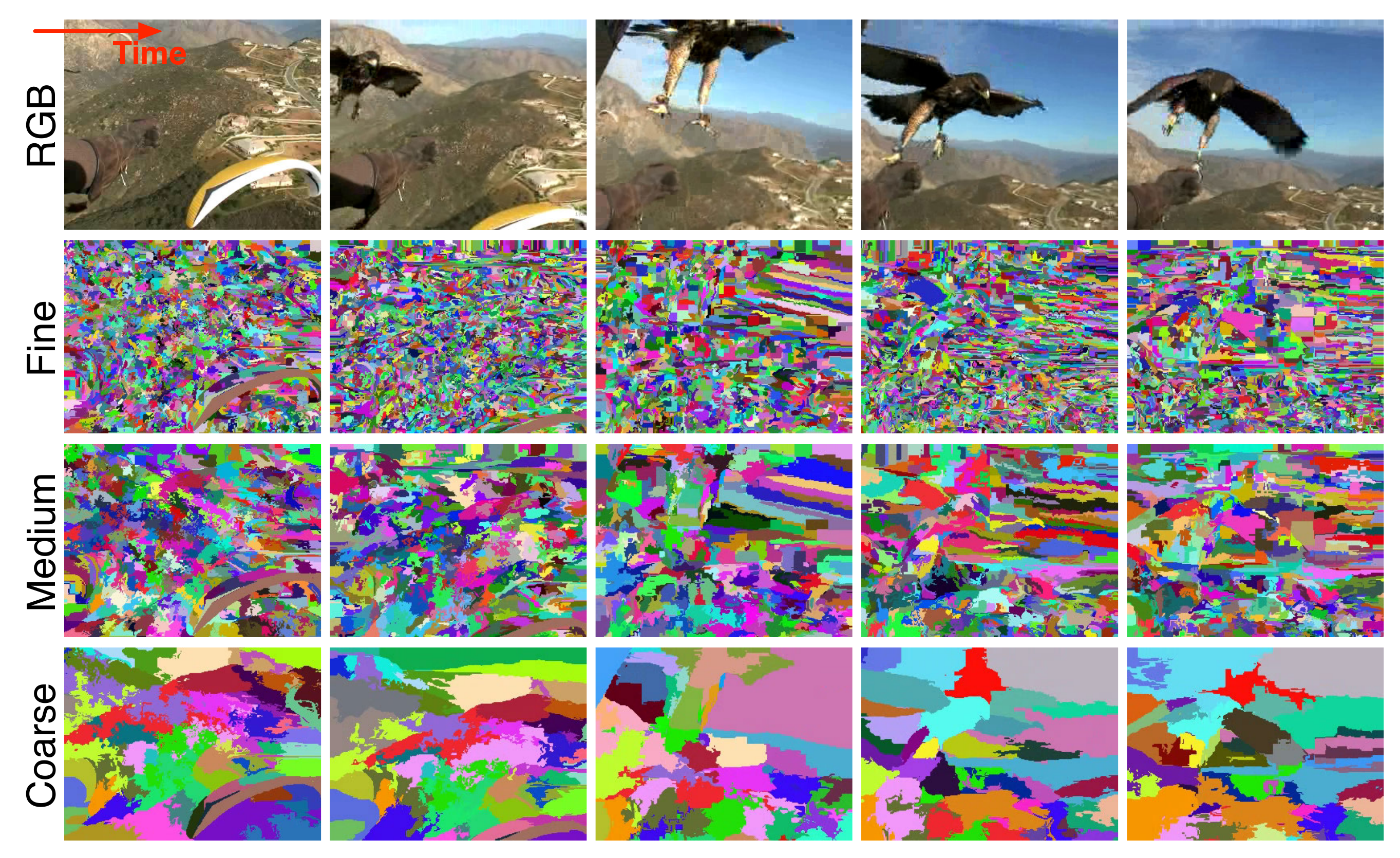} &
    \includegraphics[width=0.485\linewidth]{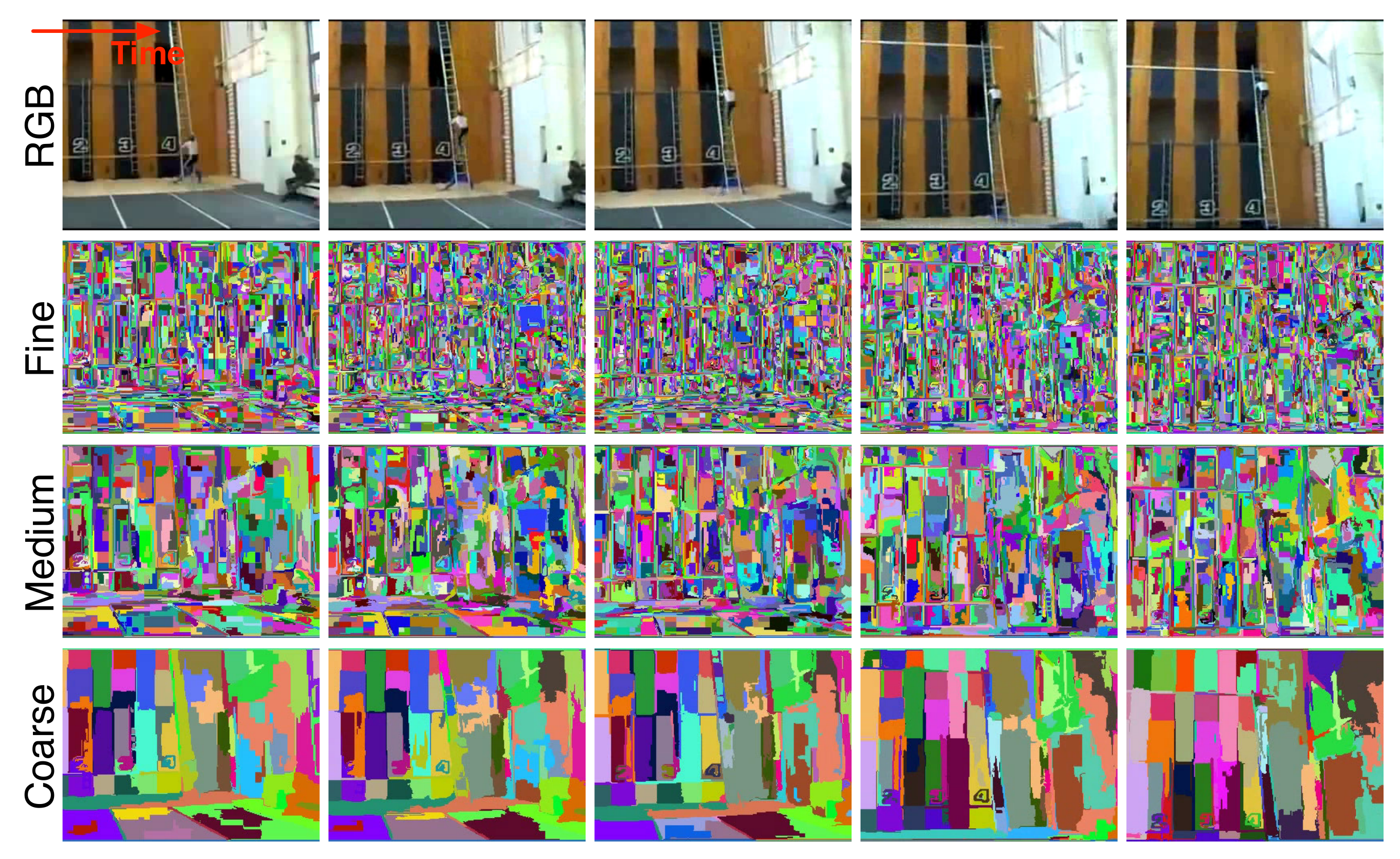}  \\
    Animal\_Flying\_Moving: 80\% Correct & Human\_Climbing\_Static: 75\% Correct \\
    \includegraphics[width=0.485\linewidth]{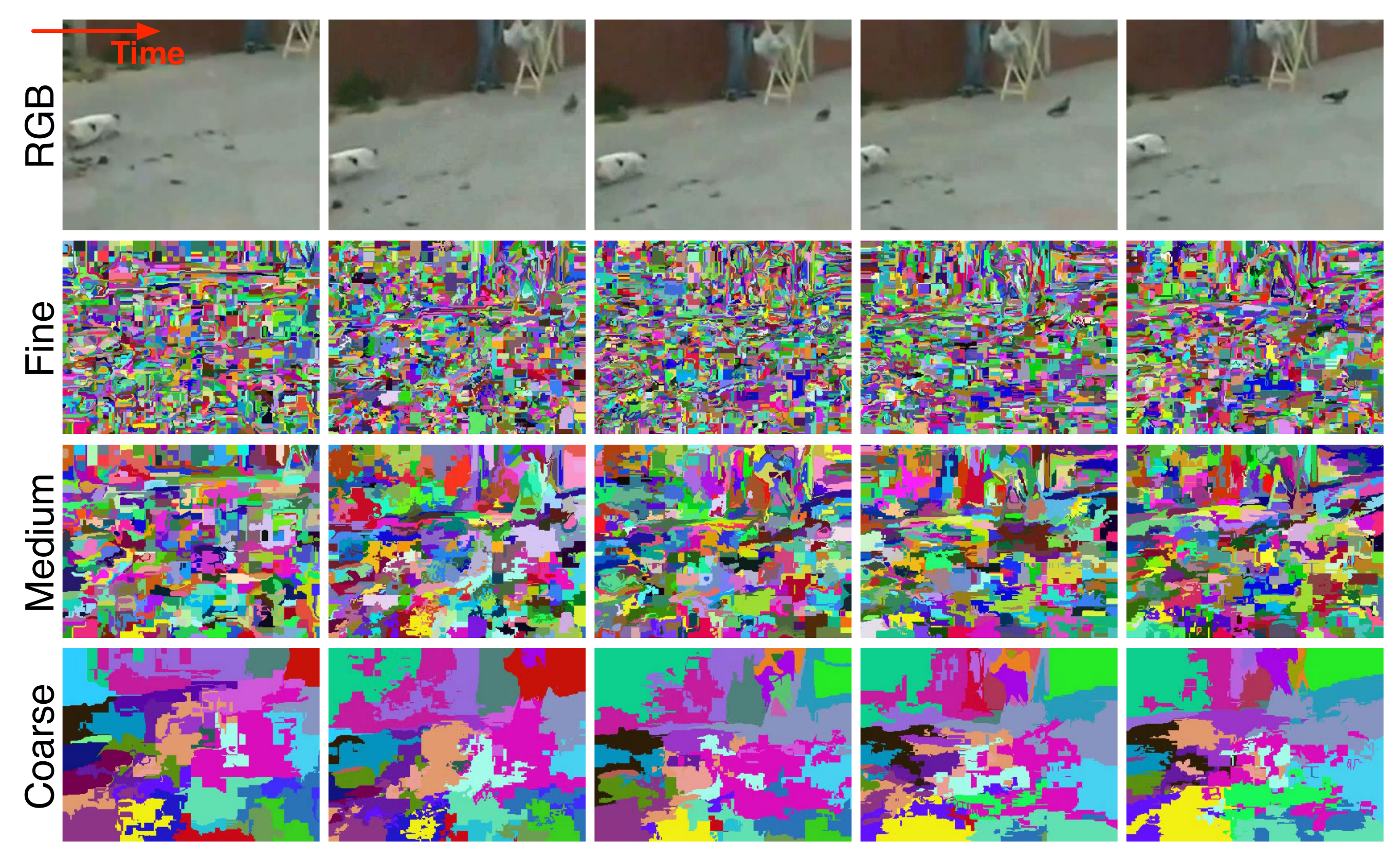} &
    \includegraphics[width=0.485\linewidth]{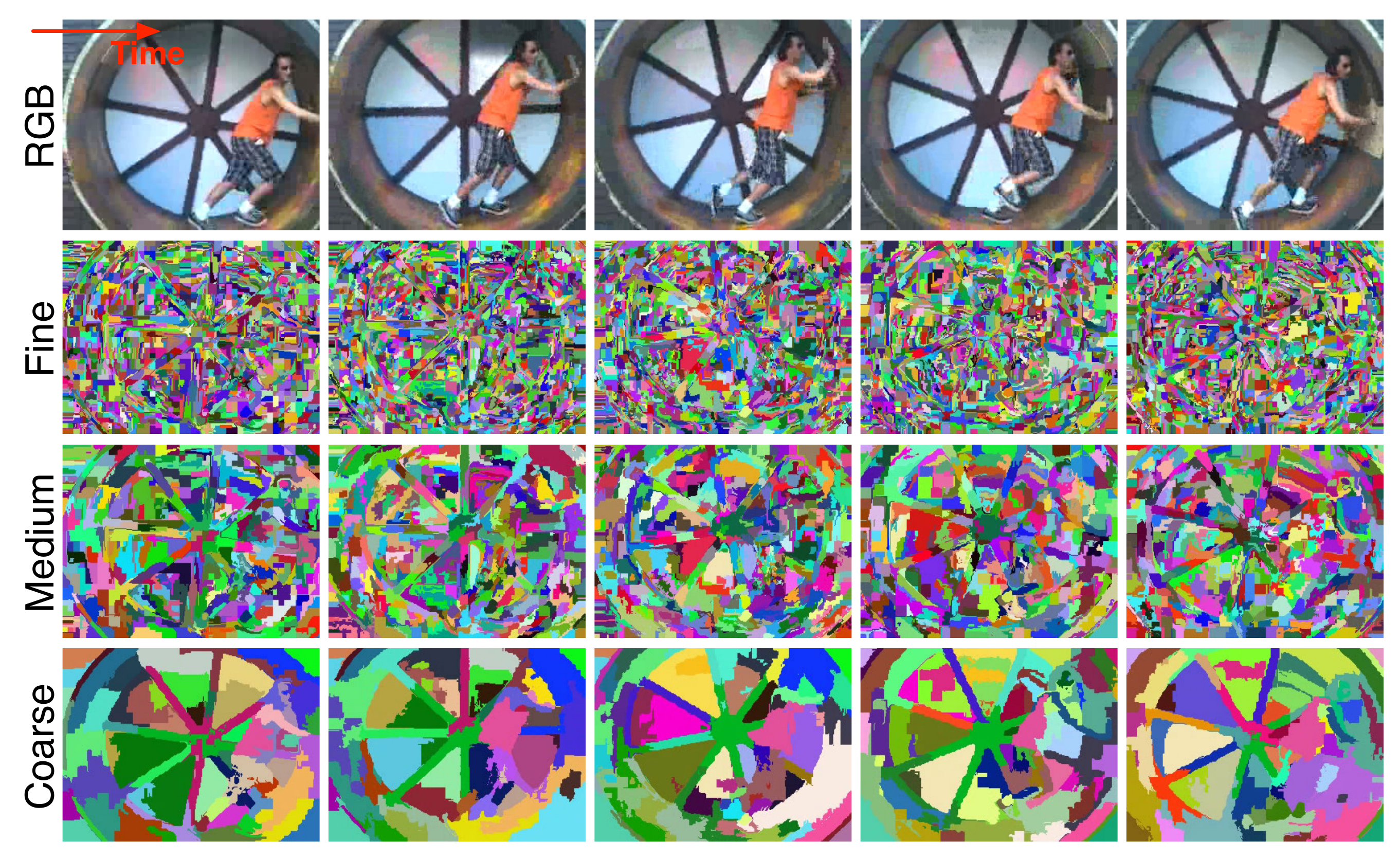}  \\
    Animal\_Crawling\_Moving: 20\% Correct & Human\_Walking\_Moving: 30\% Correct
    \end{tabular}
  \end{center}
  \caption{Visualization of videos with different levels of semantic retention. From top to bottom rows are videos picked from high, moderate, and low semantic retention resepectively. Frames are uniformly sampled from each video. We notice that supervoxel motion plays an important role in helping human observers locate the actor in a supervoxel segmentation video, which is hard to see in the montages. Therefore, we encourage people to view those videos in our project website for a better visualization.}
  \label{fig:visual}
  \vspace{-3mm}
\end{figure}

Figure \ref{fig:result_TimeByAction} shows the response time for the eight different actions. From the trend of the red curves in the figure, running and crawling get the shortest response time while the flying action takes longest. Bimodality in crawling is likely due to the very simple human baby crawling video (short response time) and very challenging cat preying video (long response time; see Figure \ref{fig:visual} for the example). The more general messages behind these results are that those unusual actions such as human flying take more time to get a response, and that those actions whose semantics have been strongly retained (resulting in higher match statistics, Table \ref{table:action_discrimination}) are generally responded to more quickly than those whose semantics have less well been retained.

%-------------------------------------------------------
\subsection{Easy, moderate and hard videos}
%-------------------------------------------------------
Finally, in Figure \ref{fig:visual}, we show montages of interesting videos, some with high action semantic retention and others with moderate or low retention. These top cases have distinct shape and motion properties that are readily transferred to the supervoxels; in the case of the running dog, the lateral motion is very strong. In the bottom left of retention examples, we see a cat crawling toward prey, but the cat is off-center from the camera and the participants likely dismiss this small off-center motion as noise for most of the video resulting in incorrect and slow responses. On the bottom right, we see a human walking in the spinning wheel. The human is walking; the wheel is spinning. There is likely semantic ambiguity here and further study is needed to understand the level and impact of the ambiguity.

%-------------------------------------------------------
\section{Machine Recognition}
\label{sec:machine}
%-------------------------------------------------------
Our human related experiment results in Section \ref{sec:results_and_analysis} indicate a strong retention of actor (82\% accuracy) and action (70\% accuracy) semantics for human observers watching the supervoxel segmentation videos. A natural follow-on question is whether machines can use the features extracted from supervoxel segmentation to perform the recognition task. Therefore, we set up an experiment to study the discriminative power of supervoxel segmentation at various hierarchy levels---applying machine learning techniques to perform video classification for actor and action, and compare it to state of the art video level features, such as dense trajectories \cite{WaKlScCVPR2011} and action bank \cite{SaCoCVPR2012}. We  conduct this experiment on the same dataset that was used previously for actor and action perception by humans.

%-------------------------------------------------------
\subsection{Supervoxel Shape Context (SSC)} 
%-------------------------------------------------------
Supervoxel segmentation contains important shape and boundary information, and the results in the visual psychophysics literature \cite{AmBiHaVR2012,GrCOiN2003,NoPhRoPerception2001,OgAiCP2007} demonstrate that such information helps higher order processes in human perception. It is our hypothesis that it is possible for machines to mimic the ability of human observers to discern actor and action by using shape related features extracted on supervoxel segmention videos. In order to reflect the shape information in supervoxel segmentation video, we adapt shape context, originally developed by Belongie et al. \cite{BeMaPuTPAMI2002}, to supervoxel segmentation video for  feature extraction. The shape context at a \textit{reference point} captures the distribution of the (supervoxel) edges relative to it, thus offering a globally discriminative characterization.

As the original shape context descriptor is intended for static images, we extend it to capture the temporal aspect of videos. First of all, we calculate the center of mass of the optical flow \cite{SuRoBlCVPR2010} for each frame in a video. This point is indicative of the greatest amount of action occurring in a frame and considered as an approximation of the actor center with respect to our single actor videos. Thus, the set of points is suitable as reference points to calculate the supervoxel shape context, which reduces the complexity required for directly determining the location in a spatiotemporal 3D video volume. The supervoxel shape context as used in this paper are the log-polar histograms of supervoxel edges calculated at the reference point on a per-frame basis. We then use five radial and 12 angular bins for the histogram quantization. Therefore, the supervoxel shape context in a video capture both spatial information by counting the locations and density of the supervoxel edges nearby the center of motion and also temporal information pertinent to action by utilizing the optical flow as an indicator of which points are most salient.

%-------------------------------------------------------
\subsection{Other Video-Based Features}
%-------------------------------------------------------
We include the state of the art video-based features for action recognition in the comparison. The span of the features include both low-level and high-level in terms of underlying semantics. Recall we visually compared the supervoxel segmentation with the other video features in Section \ref{sec:supervoxel_hierarchy}.

\noindent \textbf{Low-level Video Feature.} The low-level video features we include are dense trajectories \cite{WaKlScCVPR2011} and HOG3D \cite{KlMaScBMVC2008}. Dense trajectories are calculated from optical flow in a video and the descriptors on the trajectories capture the shape (point coordinates), appearance (histograms of oriented gradients) and motion (histograms of optical flow), as well as the differential in motion boundry (motion boundry histograms). A dense representation of trajectories guarantees a good coverage of foreground motion as well as of the surrounding context. HOG3D is the video extension of the histogram of oriented gradients based descriptors for static images \cite{DaTrCVPR2005}. It treats videos as spatio-temporal volumes and generalize the key histogram of gradients concepts to 3D and thus it captures the global shape information in a video. For both of these low-level features, we use a bag-of-words classification scheme \cite{FePeCVPR2005}.

\noindent \textbf{High-level Video Feature.} Action bank \cite{SaCoCVPR2012} is a high-level representation of video. It is comprised of many individual action detectors, such as boxing and horse back riding, that are sampled broadly in semantic space as well as viewpoint space. It computes the responses of those individual action detectors at different locations and scales in a video and ultimately transfers the video to a response vector in high-dimensional \textit{action-space}. 

Dense trajectories and action bank achieve state-of-the-art action recognition performance on popular computer vision datasets, such as UCF50 \cite{ReShMVAP2012}.

%-------------------------------------------------------
\subsection{Experiment and Analysis} 
%-------------------------------------------------------
We set up the experiment to classify videos based on actor and action 
separately. For actor, it is a two class (human or animal) classification task. 
For action, it is \textit{one versus all} for each of the eight actions. For 
example, when we train a classifier for action \textit{walking}, we use videos 
containing walking as positive samples and the rest videos containing the other 
seven actions as negative samples. Due to the small size of our experiment data 
set, the classification is done with a \textit{leave-one-out} setup.

For supervoxel shape context (SSC), we again extract the features at different levels in a hierarchy: fine, median and coarse segmentations, which are the same as in human experiment. For dense trajectories (DT) and HOG3D, we use a codebook of 20 words, which is small but necessary due to the data set size and empirically found to outperform other sized codebooks. We also note that all the action detectors in action bank are from human-based templates, which limits its performance in actor classification. We refer to \cite{SaCoCVPR2012} for a complete list of action templates.

\begin{table}[!t]
  \centering
  \includegraphics[width=\textwidth]{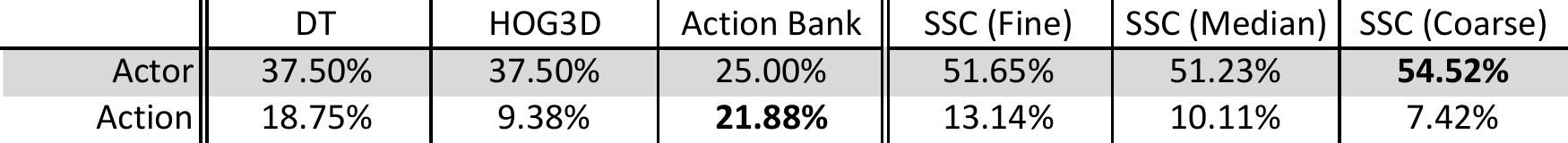}
  \caption{Overall machine classification accuracy. The leading score of each task is highlighted in bold font.}
  \label{tab:machine_average}
\end{table}

\noindent \textbf{Overall Performance.} Table \ref{tab:machine_average} shows the overall machine classification accuracy. Surprisingly, for actor classification, SSC gives much higher accuracy than the others and SSC at coarse level achieves the best performance. We suspect the reason why the performance of other features is worse than chance (50\%) is because of a lack of discrimination in the feature (e.g., action bank has only human-based templates), or a lack of sufficient training samples in the data set. We also suspect that the shape information at coarse levels is cleaner than at finer levels for training a actor classifier. For action classification, action bank achieves the leading performance, which is probably due to the underlying high-level semantics. Interestingly, SSC at finer levels perform better than at coarser levels. Note that the action classification by chance is 12.5\%.

\begin{figure}[!t]
  \centering
  \includegraphics[width=0.85\textwidth]{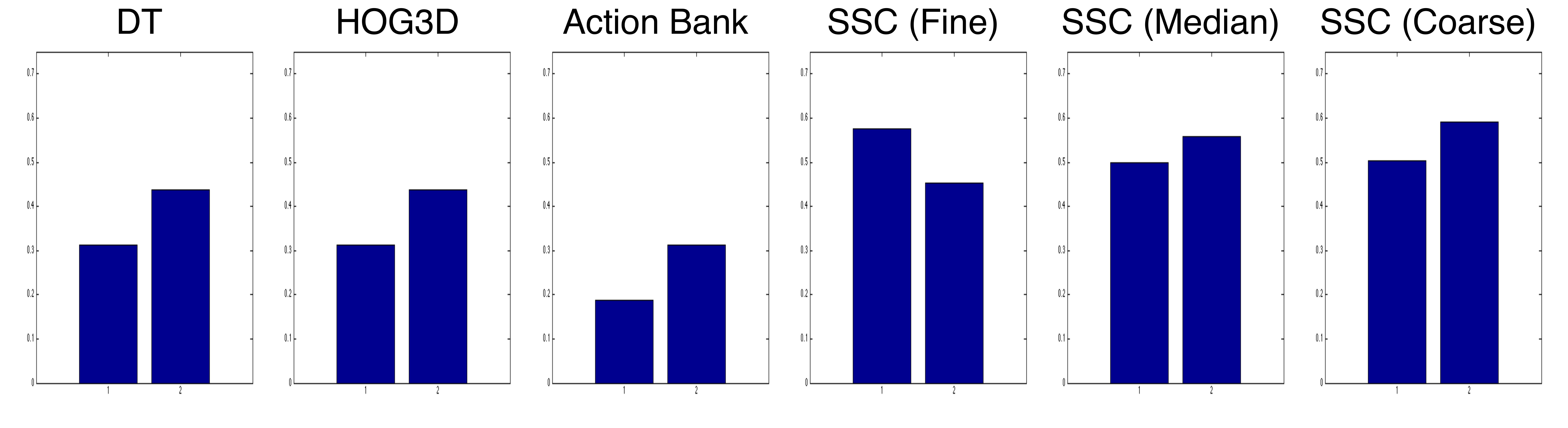}
  \caption{Classification accuracy on two class actors. In each sub-bar chart, left side bar is \textit{animal} and right side bar is \textit{human}.}
  \label{fig:machine_actor}
\end{figure}

\noindent \textbf{Actor.} Figure \ref{fig:machine_actor} shows the 
classification performance on \textit{animal} and \textit{human} respectively by 
using different types of features. Overall, the classification on human has 
higher accuracy than on animal, which is similar to human perception, as in Section \ref{subsubsec:human_actor}.

\begin{figure}[!t]
  \centering
  \includegraphics[width=0.85\textwidth]{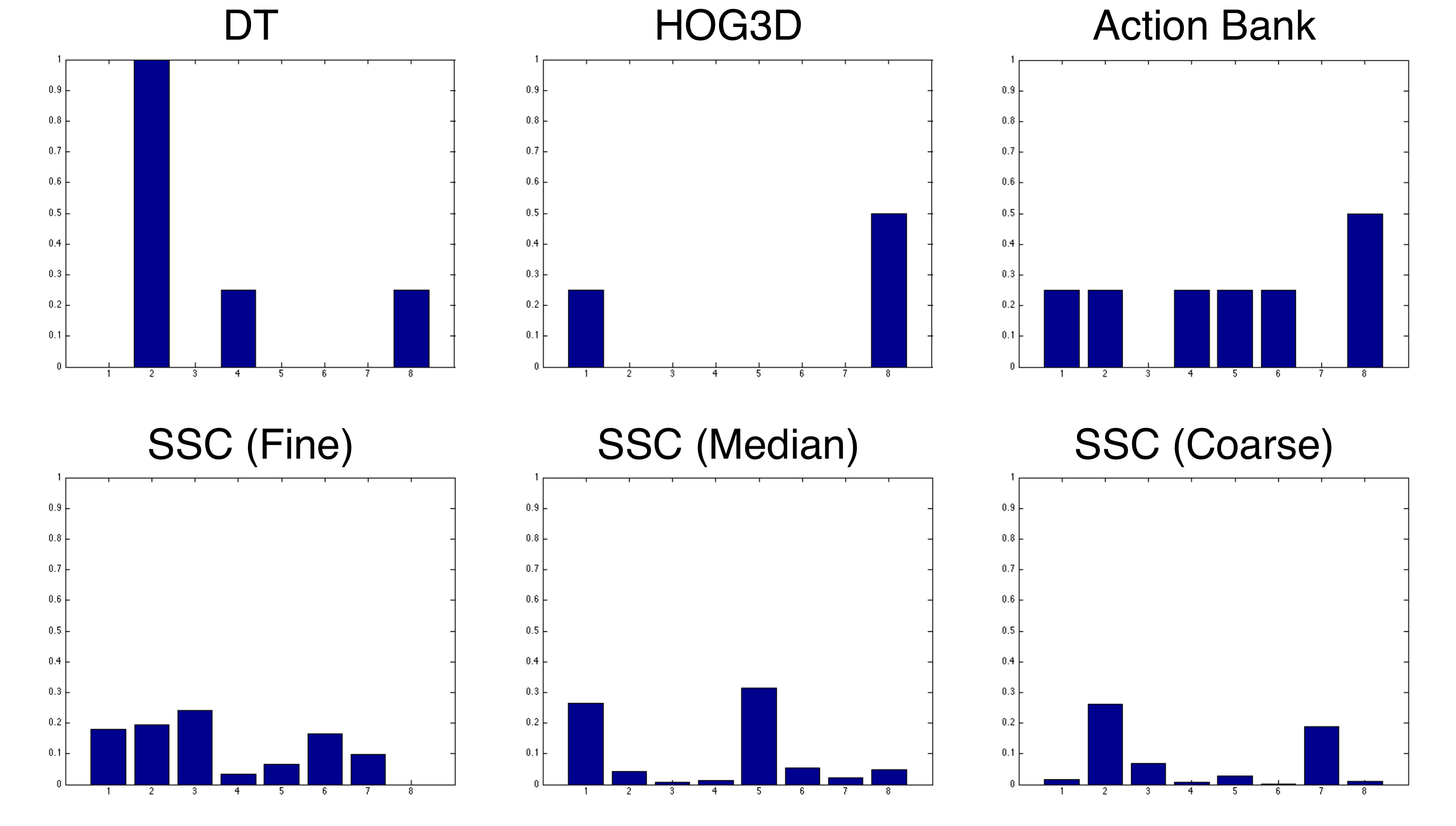}
  \caption{Classification accuracy on eight actions. Bars from left to right represent \textit{eating}, \textit{climbing}, \textit{walking}, \textit{jumping}, \textit{crawling}, \textit{spinning}, \textit{flying} and \textit{running}.}
  \label{fig:machine_action}
\end{figure}

\begin{figure}[!t]
  \centering
  \includegraphics[width=0.75\textwidth]{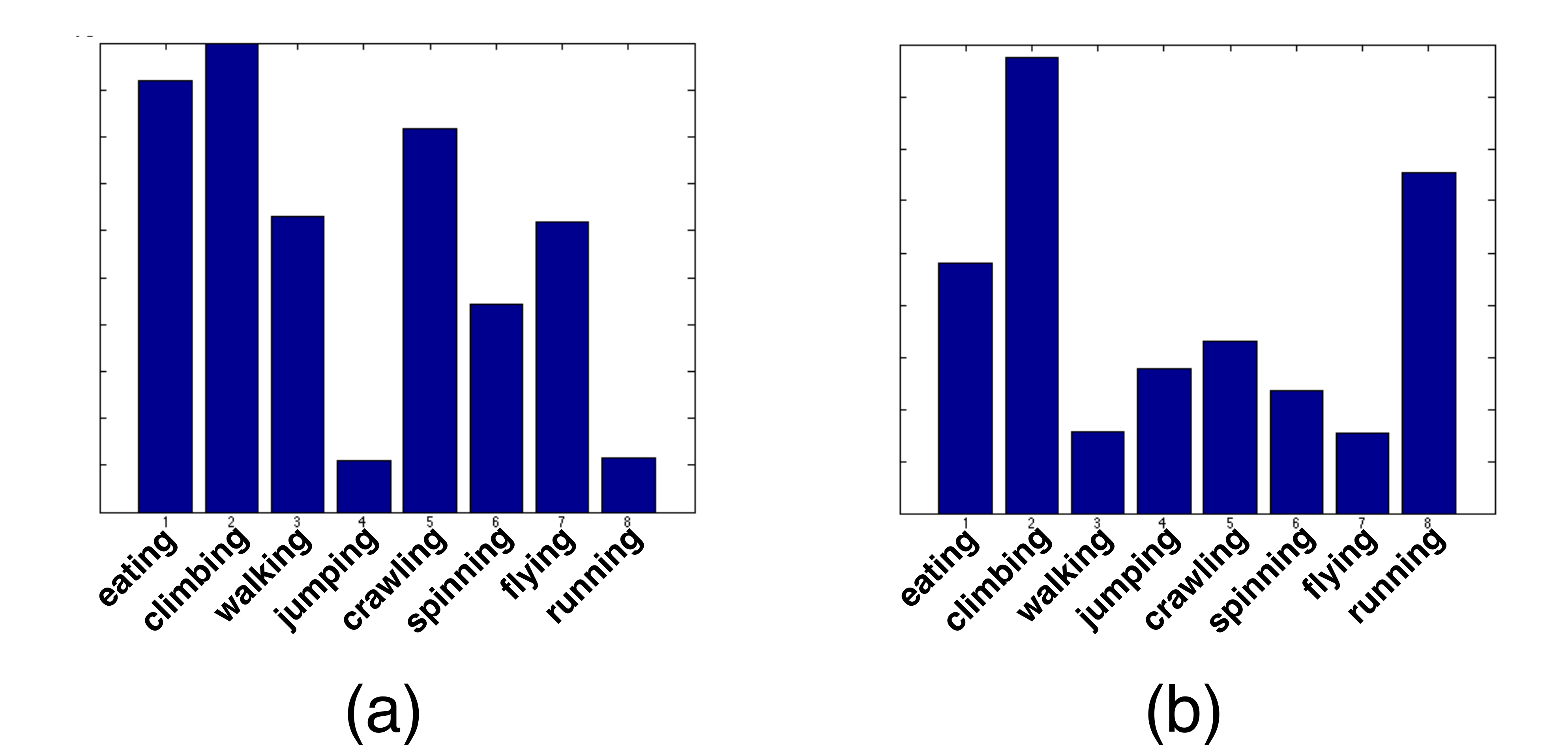}
  \caption{The summed accuracy on eight actions across features. (a) summed action accuracy of SSC at all different levels; (b) summed action accuracy across all types of features.}
  \label{fig:machine_action_all}
\end{figure}

\noindent \textbf{Action.} Figure \ref{fig:machine_action} shows the 
classification performance on eight actions. Different features have very 
different performance on actions. This is expected since the different types of 
feature have very distinct foci in a video. To compare with human perception of 
actions in supervoxel segmentation, we are also interested to know the summed 
accuracy on actions across supervoxel shape context at different levels, as 
shown in Figure \ref{fig:machine_action_all} (a). Suprisingly, the machine 
results on action recognition agree with human perceptions (Section 
\ref{subsubsec:human_action}) to a large degree. The top scoring actions of  
human perceptions (\textit{climbing} and \textit{eating}), are also the top 
performing ones of machine recognition. Furthermore, the lowest scoring ones from human perception (\textit{walking}, \textit{jumping}, and \textit{spinning}) are also among the worst ones of machine recognition. The summed accuracy on actions across all types of features, as shown in Figure \ref{fig:machine_action_all} (b), further enforces this finding.

\noindent \textbf{Summary.} Our experimental results demonstrate that the rich 
semantics in supervoxel segmentation can be well used in machine recognition to 
achieve competitive results when comparing with state of the art video level 
features. The overall machine recognition of actor and action seems to follow 
the same trend as that of human perception but more work needs to be done to get the machine recognition models up to par with the humans in terms of recognition performance.

%-------------------------------------------------------
%-------------------------------------------------------
\section{Conclusion}
\label{sec:conclusion}
%-------------------------------------------------------
%-------------------------------------------------------
In this paper, we explore the degree to which actor and action semantics are 
retained in video supervoxel segmentation. We design and conduct a systematic 
study to answer a set of questions related to this semantic retention. Our 
experiment results indicate strong retention of actor and action semantics: 
human perception achieves 82\% accuracy on actor and 70\% on action. 
Furthermore, this underlying semantics in supervoxel segmentation can be well 
used in machine recognition tasks. The supervoxel shape context achieves 
competitive results when comparing with state of the art video level features, 
but it is still far away from human performance. Our overall findings suggest 
that supervoxel segmentation is a rich decomposition of the video content, 
compressing the signal significantly while retaining enough semantic information 
to remain discriminative.  The next challenge is to develop better methods for representation of the supervoxels for machine modeling.

% The actor and action perception experiment we have reported is in a closed-world setting. In the future, we will explore the open world problem: how much semantic information is retained in supervoxel segmentation. We will let participants use lingual description to describe the supervoxel human perception (such as one can recover the woman in the painting hanging on the wall in Figure \ref{fig:hieseg}). We will then compare this human perception to a machine perception using our recent video-to-text engine \cite{DaXuDoCVPR2013}. We will also collect a larger number of videos and conduct experiments with more participants to overcome the limitation of 20 participants and 32 input videos.

%-------------------------------------------------------
%-------------------------------------------------------
\section*{Acknowledgements}
%-------------------------------------------------------
%-------------------------------------------------------
CX, RFD, and JJC are partially supported by NSF CAREER IIS-0845282 and DARPA Mind's Eye W911NF-10-2-0062.  SJH and CH are partially supported by the McDonnell Foundation.  The authors would like to thank the anonymous participants as well as Cody Boppert and Henry Keane for their help in gathering the data set.

%-------------------------------------------------------
%-------------------------------------------------------
{\small
\bibliographystyle{IEEEtran}
\bibliography{IEEEabrv,arXiv_SvxHuman}
}
%-------------------------------------------------------
%-------------------------------------------------------
\end{document}